%% file: main.tex
\documentclass[11pt]{article}

\usepackage{setspace}
\usepackage[showframe=false, margin=1.0in]{geometry}

\usepackage[utf8]{inputenc}
\usepackage{wrapfig}
\usepackage[lofdepth,lotdepth]{subfig}
\usepackage{booktabs}
\usepackage{rotating}
\usepackage{caption}
\usepackage{csquotes}
\usepackage{stfloats}
\usepackage[T1]{fontenc}
\usepackage{textcomp}
\usepackage{times}
\usepackage{tabularx}
\usepackage{pdfpages}

\usepackage{framed} %um Boxen zu machen

\usepackage{color}
\definecolor{black}{gray}{0} % 10% gray

\usepackage{tabularx}
\newcolumntype{b}{X}
\newcolumntype{s}{>{\hsize=.5\hsize}X}

\newcommand{\score}[1]{\psi(#1)}

\newcommand{\SupFig}{Supplementary Fig.\xspace}
\newcommand{\SupTab}{Supplementary Table}
\newcommand{\SupTabs}{Supplementary Tables}

\title{United States Politicians' Tone Became More Negative\\with 2016 Primary Campaigns}

\author{%
Jonathan K\"ulz$^{1}$,
Andreas Spitz$^{2}$,
Ahmad Abu-Akel$^{3}$,
Stephan G\"unnemann$^{1}$,
Robert West$^{4}$\footnote{To whom correspondence should be addressed: robert.west@epfl.ch}\\
{\small $^{1}$Department of Informatics, Technical University of Munich, Germany}\\
{\small $^{2}$Department of Computer and Information Science, University of Konstanz, Germany}\\
{\small $^{3}$School of Psychological Sciences, University of Haifa, Israel}\\
{\small $^{4}$School of Computer and Communication Sciences, Ecole Polytechnique F\'ed\'erale de Lausanne, Switzerland}
}

\date{}

\usepackage[authordate,
			backend=biber,
            maxbibnames=99,
			language=auto,
			url=true,
			doi= true,
			isbn=false]{biblatex-chicago}

\makeatletter
\AtEveryBibitem{%
  \global\undef\bbx@lasthash%
  \clearfield{extrayear}}
\makeatother

\input{dlab_macros}

\begin{document}

\maketitle

% Updated Numbers (old ones were based on thesis submission in October, they SHOULD be all correct in the draft now):
% - Nr Politicians that actually uttered a quote that was analysed: 18,627
% - Nr Quotes Analysed (incl. multi-attributions): 26,034,267
% - Nr unique Quotes Analysed: 24,267,018 → The one I put in the paper

% \renewcommand{\abstractname}{Summary}
\begin{abstract}
\noindent
There is a widespread belief that the tone of US political language has become more negative recently, in particular when Donald Trump entered politics.
At the same time, there is disagreement as to whether Trump changed or merely continued previous trends.
To date, data-driven evidence regarding these questions is scarce, partly due to the difficulty of obtaining a comprehensive, longitudinal record of politicians' utterances.
Here we apply psycholinguistic tools to a novel, comprehensive corpus of 24 million quotes from online news attributed to 18,627 US politicians in order to analyze how the tone of US politicians' language evolved between 2008 and 2020.
We show that, whereas the frequency of negative emotion words had decreased continuously during Obama's tenure, it suddenly and lastingly increased with the 2016 primary campaigns, by 1.6 pre-campaign standard deviations, or 8\% of the pre-campaign mean, in a pattern that emerges across parties.
The effect size drops by 40\% when omitting Trump's quotes, and by 50\% when averaging over speakers rather than quotes, implying that prominent speakers, and Trump in particular, have disproportionately, though not exclusively, contributed to the rise in negative language.
This work provides the first large-scale data-driven evidence of a drastic shift toward a more negative political tone following Trump's campaign start as a catalyst, with important implications for the debate about the state of US politics.
\end{abstract}

%%%%%%%%%%%%%%%%%%%%%%%%%%%%%%%% START %%%%%%%%%%%%%%%%%%%%%%%%%%%%%%%% 

\noindent
A vast majority of Americans---85\% in a representative survey by the \cite{pew2019discourse}---have the impression that ``the tone and nature of political debate in the United States has become more negative in recent years''. Many see a cause in Donald Trump, who a majority (55\%) think ``has changed the tone and nature of political debate [\dots{}]\ for the worse'', whereas only 24\% think he ``has changed it for the better''~\autocite{pew2019discourse}.
The purpose of the present article is to investigate whether these subjective impressions reflect the true state of US political discourse.
The answer to this question comes with tangible societal implications:
as politics impacts nearly every aspect of our personal lives~\autocite{Smith2019cost}, changes in political climate can directly affect not only politics itself, but also the well-being of all citizens.

Although longitudinally conducted voter surveys, notably the American National Election Studies, have shown that negative affect towards members of the other party \autocite{abramowitz2016partisanship, iyengar2012affect, iyengar2019affec}, polarization \autocite{abramowitz2008polarization}, and partisan voting \autocite{bafumi2009partisan} have steadily increased over the last decades, such survey-based studies do not directly measure the tone of political debate---a linguistic phenomenon---and thus cannot answer whether Americans' impression of increased negativity is accurate.
In contrast, we analyze US politicians' language directly, in an objective, data-driven manner, asking:
First, is it true that US politicians' tone has become more negative in recent years?
Second, if so, did Donald Trump's entering the political arena bring about an abrupt shift \autocite{costello2016trump,crandall2018changing}, or did it merely continue a previously existing trend \autocite{fritze2021trump,james2021effects}?

% "Trump took advantage of preexisting distrust, polarization and frustration in America and used rhetorical strategies that were designed to make all of those negative things worse," said Jennifer Mercieca, a historian of American political rhetoric and an associate professor at Texas A&M University. \autocite{fritze2021trump}

Our methodology draws on a long history of research on the language of politics and its function in democracies \autocite{dunmire2012pda, farelly2010discourse, tenorio2002minister, zheng2000rhetoric}. For instance, prior work has used records of spoken and written political language
to establish the prevalence of negative language among political extremists \autocite{frimer2019extremists};
to quantify growing partisanship and polarization \autocite{gentzkow2019partisanship}, as well as displayed happiness \autocite{wojcik2015hapiness}, among US Congress members;
to analyze political leaders' psychological attributes such as certainty and analytical thinking \autocite{jordan2019leader};
to quantify the turbulence of Trump's presidency \autocite{dodds2021computational};
or to measure the effect of linguistic features on the success of US presidential candidates \autocite{thoemmes2007complexity} and on public approval of US Congress \autocite{frimer2016congress}.
A combination of political discourse analysis and psychological measurement tools has further been applied to obtain insights into the personality traits and sentiments of politicians in general \autocite{kangas2014software, tumasjan2010elections}, as well Donald Trump in particular \autocite{kreis2017tweets, ahmadian2017trump, depryck2017science, lewandowsky2020using}.

One limiting factor in the above-cited works is the representativeness and completeness of the underlying data, since subtle social or political behavior may only reveal itself in sufficiently big and rigorously processed data \autocite{monroe2015no, lazer2021meaningful}. On the one hand, Congressional records \autocite{gentzkow2019partisanship, wojcik2015hapiness, frimer2016congress} and transcripts of public speeches \autocite{thoemmes2007complexity, jordan2019leader} record scarce events and do not mirror political discourse as perceived by the average American, whose subjective impression of growing negativity we aim to compare with objective measurements of politicians' language. On the other hand, news text, despite being a better proxy for the average American's exposure to political discourse, for the most part does not capture politicians' utterances directly, but largely reports events (laced with occasional direct quotes) and frequently focuses on a commentator's perspective. Additionally, most of the above-cited linguistic analyses are not longitudinal in nature, but focus on specific points in time.

Transcending these shortcomings, we take a novel approach leveraging Quotebank \autocite{vaucher2021quotebank}, a recently released corpus of nearly a quarter-billion (235 million) unique quotes extracted from 127 million online news articles published by a comprehensive 
% \autocite{west2021postmortem}
set of online news sources over the course of nearly 12 years (September 2008 to April 2020) and automatically attributed to the speakers who likely uttered them by a machine learning algorithm. By focusing on US politicians, we derived a subset of 24 million quotes by 18,627 speakers, enriched with biographic information from the Wikidata knowledge base \autocite{wikidata}.
(Details about data in \textit{Materials and Methods}.)
As no comparable dataset of speaker-attributed quotes was available before, Quotebank enables us to analyze the tone of US politicians' public language (as seen through the lens of online news media) at a level of representativeness and completeness that was previously impossible, without confounding politicians' direct utterances with the surrounding news text.

\begin{figure*}[htb]
	\includegraphics[width=\linewidth]{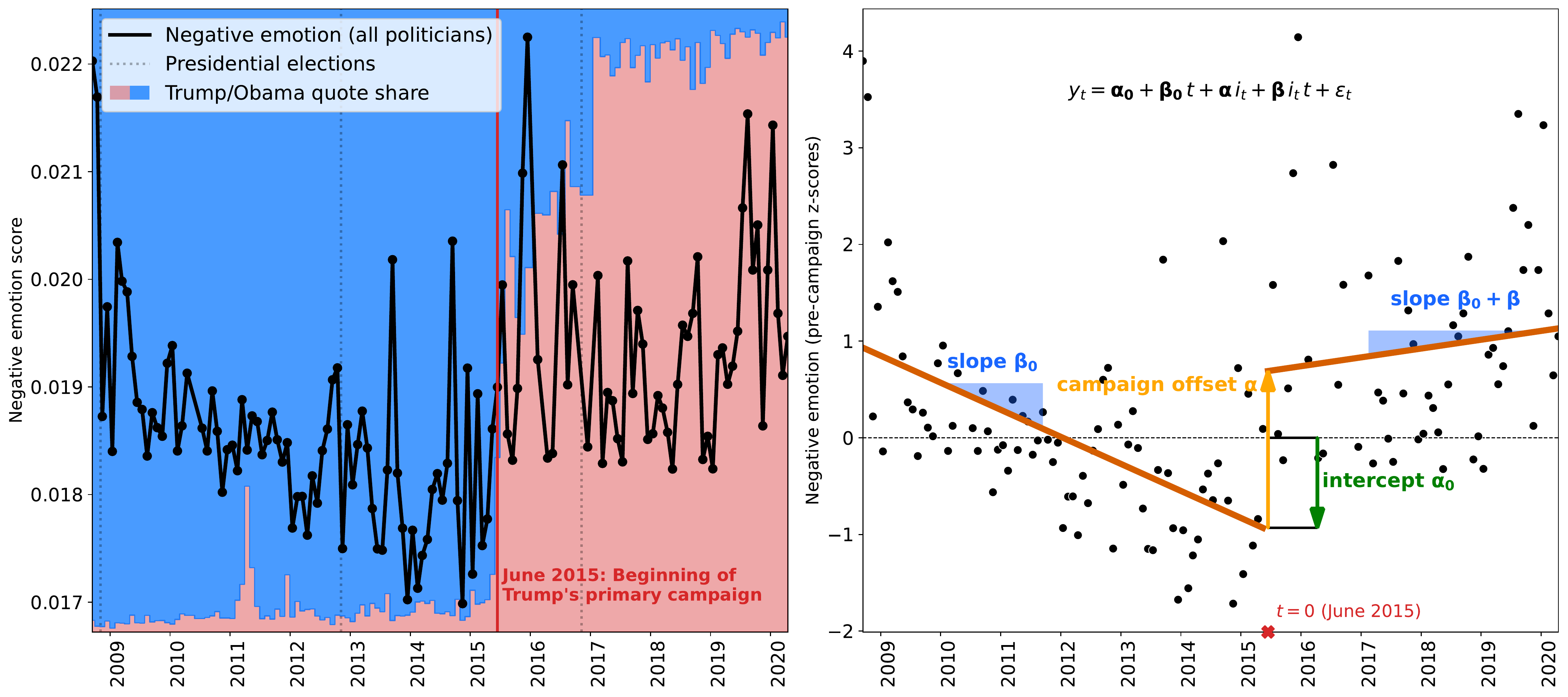}

	\hspace{0.26\linewidth} (a) \hspace{0.46\linewidth} (b)
	\caption{
	\textbf{Quantifying the evolution of negative language in US politics (2008--2020).}
	\textbf{(a)}~The black points show the fraction of negative emotion words, averaged monthly over all quotes from all 18,627 quoted politicians. The red \vs\ blue background shows the quote share of Trump \vs\ Obama (if Trump had $T$ quotes and Obama had $O$ quotes in a given month, the respective red bar covers a fraction $T/(T+O)$ of the full $y$-range).
	Whereas the frequency of negative emotion words had decreased continuously during the first 6.5 years of Obama's tenure, it suddenly and lastingly increased in June 2015, when Trump's primary campaign started and his quote share began to surpass Obama's.
	\textbf{(b)}~Regression analysis: The black points again show the fraction of negative emotion words, but now as $z$-scores (\ie, after subtracting the pre-campaign mean and dividing by the pre-campaign standard deviation).
	In red, we plot regression lines for the periods before and after June 2015.
	The coefficients of the ordinary least squares regression model
	$y_t = \alpha_0
	+ \beta_0 \,t
	+ \alpha \,i_{t}
	+ \beta \,i_{t} \,t
	+ \varepsilon_{t}$
	(where $t$ is the number of months since June 2015, and $i_t$ indicates whether $t \geq 0$; \cf\ \Eqnref{eqn:regression})
	quantify the slopes of both lines, as well as the sudden increase of $\alpha=1.6$ pre-campaign standard deviations coinciding with the discontinuity in June 2015 ($t=0$), as visualized.
	}
	\label{fig:fig0}
\end{figure*}

In order to quantify the prevalence of negative language over time, we use established psycholinguistic tools~\autocite{pennebaker2007liwc} to score each quote with respect to its emotional content,
% pool quotes across politicians,
aggregate quotes by month, and work with the resulting time series. Anchored in the average American's subjective perception that Donald Trump has changed the tone and nature of political debate (see above), we hypothesized the start of his primary campaign in June 2015 as an incision point and fitted linear regression models \autocite{hausman2018regression} with a discontinuity in June 2015 to the time series, as illustrated in \Figref{fig:fig0}.

The results (\cf\ \Figref{fig:fig0}) provide clear evidence of a sudden shift coinciding with the hypothesized discontinuity at the start of Trump's primary campaign in June 2015, when the overall political tone became abruptly more negative. The effect was large and highly significant: the frequency of negative emotion words soared by 1.6 pre-campaign standard deviations, or by 8\% of the pre-campaign mean. Moreover, whereas negative language had decreased continuously during (at least) the first 6.5 years of Obama's tenure, this trend came to a halt with the June 2015 discontinuity, and the level of negative language remained high afterwards. Similar effects were observed for specific subtypes of negative language (anger, anxiety, and sadness), as well as for swearing terms.
Qualitatively, the patterns are universal, emerging within each party and within strata of speaker prominence. Quantitatively, the language of Democrats, more prominent speakers, Congress members, and members of the opposition party is, \textit{ceteris paribus,} overall more negative, but the shift at the discontinuity remains (and more strongly so for Republicans) when adjusting for these biographic attributes.

These population-level effects are not the results of systematic shifts in the distribution of quoted politicians, but are mirrored at the individual level for a majority of the most-quoted politicians. Moreover, the effect size drops by 40\% when omitting Donald Trump's quotes from the analysis, and by 50\% when weighing each speaker equally, rather than by the number of their quoted utterances (but the effect remains highly significant in both cases). Both findings imply that prominent speakers---and particularly Donald Trump---have disproportionately, though not exclusively, contributed to the rise in negative language.

Taken together, these results objectively confirm the subjective impression held by most Americans~\autocite{pew2019discourse}: recent years have indeed seen a profound and lasting change toward a more negative tone in US politicians' language as reflected in online news, with the 2016 primary campaigns acting as a turning point.
Moreover, contrary to some commentators' assessment \autocite{fritze2021trump,james2021effects}, Donald Trump's appearance in the political arena was linked to a directional change, rather than a continuation of previously existing trends in political tone.
Whether these effects are fully driven by changes in politicians' behavior or whether they are exacerbated by a shifting selection bias on behalf of the media remains an important open question (see \textit{Discussion}).
Either way, the results presented here have implications for how we see both the past and the future of US politics.
Regarding the past, they emphasize the symptoms of growing toxicity in US politics from a new angle.
Regarding the future, they highlight the danger of a positive feedback loop of negativity.

%%%%%%%%%%%%%%%%%%%%%%%%%%%%%%%%%%%%%%%%%%%%%%%%%%%%%%%%%%%%
%%%%%%%%%%%%%%%%%%%%%%%%%%%%%%%%%%%%%%%%%%%%%%%%%%%%%%%%%%%%

%%%%%%%%%%%%%%%%%%%%%%%%%%%%%%%%%%%%%%%%%%%%%%%%%%%%%%%%%%%%
%%%%%%%%%%%%%%%%%%%%%%%%%%%%%%%%%%%%%%%%%%%%%%%%%%%%%%%%%%%%
%%%%%%%%%%%%%%%%%%%%%%%%%%%%%%%%%%%%%%%%%%%%%%%%%%%%%%%%%%%%

\section*{Results}

% Useful LIWC reference from 2013: https://aclanthology.org/S13-1042.pdf 
% "Choosing the Right Words: Characterizing and Reducing Error of the Word Count Approach"

To quantify negative language in quotes, we used Linguistic Inquiry and Word Count (LIWC) \autocite{pennebaker2007liwc}, which provides a dictionary of words belonging to various
linguistic, % (e.g., first-person singular pronouns, conjunctions),
psychological, % (e.g., anger, achievement),
and topical % (e.g., leisure, money)
categories,
and whose validity has been established by numerous studies in different contexts and domains \autocite{alpers2005cancer, kahn2007emotions, settanni2015facebook}.
We computed a negative-emotion score for each of the 24 million quotes via the percentage of the quote's constituent words that belong to LIWC's \textit{negative emotion} category.
Analogously, we computed scores for LIWC's three subcategories of the \textit{negative emotion} category---\textit{anger,} \textit{anxiety,} and \textit{sadness}---as well as for the \textit{swear words} category.
We collectively refer to these five categories as ``negative language''.%
\footnote{%
As a robustness check, we replicated the analysis using the dictionary provided by Empath \autocite{fast2016empath}
% (categories: \textit{negative emotion,} \textit{positive emotion,} \textit{swearing})
instead of LIWC.
In contrast to LIWC, the categories in Empath's dictionary were not hand-crafted but generated by a deep-learning model based on a small set of seed words.
The results, which are consistent with those from the LIWC-based analysis, are provided in \SupFig{} S1, S2, S3, and S4, and in \SupTab{} S1.}

In order to obtain a time series for each word category, we averaged individual quote scores by month.
By giving each quote the same weight when averaging, we obtain \textit{quote-level aggregates;}
by giving each speaker the same weight, \textit{speaker-level aggregates.}
(Formal definitions in \textit{Materials and Methods}.)
%%% we could say that we'd ideally weight quotes by how often they were read, but we don't have this info
Quote-level aggregates give more weight to more frequently quoted politicians, and thus better capture the overall tone of political language as reflected in the news.
% through the eyes of a media consumer reading online news articles. This way, prominent politicians naturally dominate the quote-level aggregate scores.
Whenever, on the contrary, we reason about politicians, rather than about the overall media climate created by all politicians' joint output, we use speaker-level aggregates as the more appropriate aggregation.

\Tabref{tab:scores} summarizes the prevalence of the above five word categories via the means and standard deviations of the respective quote-level aggregates during the pre-campaign period (September 2008 through May 2015).
We observe that
one in 54 words expresses a negative emotion;
one in 155, anger;
one in 339, anxiety;
and one in 285, sadness.
Swear words, at a rate of one in 2,329, are least common.
% , whereas positive emotion words, at a rate of one in 25, are most common.
Given this wide range of frequencies, and in order to make effect sizes comparable across word categories, we standardize all monthly scores category-wise by subtracting the respective pre-campaign mean and dividing by the respective pre-campaign standard deviation.%
\footnote{%
In order to facilitate the comparability and interpretability of results, standardization always uses the means and standard deviations computed on quote-level aggregates involving all speakers, even in analyses of speaker-level aggregates or of quote-level aggregates when omitting individual speakers.}
The resulting effect sizes, in units of pre-campaign standard deviations, become more palpable when expressed as multiples of the corresponding pre-campaign means.
For this purpose, the rightmost column of \Tabref{tab:scores} lists coefficients of variation, \ie, ratios of standard deviations and means.

%%%%%%%%%%%%%%%%%%%%%%%%%%%%%%%%%%%%%%%%%%%%%%%%%%%%%%%%%%%%
\begin{table}
\centering
\caption{
\textbf{Overall statistics of LIWC negative-language word categories.}
Means $\mu$ and standard deviations $\sigma$ were calculated over monthly quote-level aggregates from the pre-campaign period (September 2008 through May 2015).
The value $n=1/\mu$ in the fourth column implies that, in an average quote, on average every $n$-th word belongs to the respective category.
The coefficients of variation, $\sigma/\mu$, shown in the fifth column, allow to easily translate pre-campaign standard deviations (as shown on the $y$-axes of time series plots) into fractions of pre-campaign means.
The most frequent words per category are listed in \SupTabs{} S2 and S3.
}
\label{tab:scores}
\begin{tabular}{lrrrr}
    \toprule
    Word category     &    Mean $\mu$ & Std.\ dev.\ $\sigma$ & $1/\mu$ & $\sigma/\mu$ \\
    \midrule
    Negative emotion &  0.01853 &  0.00090 &      54 &          4.8\% \\
    Anger            &  0.00647 &  0.00055 &     155 &          8.4\% \\
    Anxiety          &  0.00295 &  0.00022 &     339 &          7.3\% \\
    Sadness          &  0.00351 &  0.00014 &     285 &          4.1\% \\
    Swear words      &  0.00043 &  0.00015 &    2,329 &         33.9\% \\
    %%% Before rounding:
    % Negative emotion &  0.01853 &  0.00090 &      54.0 &          4.8\% \\
    % Anger            &  0.00647 &  0.00055 &     154.6 &          8.4\% \\
    % Anxiety          &  0.00295 &  0.00022 &     339.4 &          7.3\% \\
    % Sadness          &  0.00351 &  0.00014 &     284.8 &          4.1\% \\
    % Swear words      &  0.00043 &  0.00015 &    2328.9 &         33.9\% \\
    \bottomrule
\end{tabular}
\end{table}
%%%%%%%%%%%%%%%%%%%%%%%%%%%%%%%%%%%%%%%%%%%%%%%%%%%%%%%%%%%%

%%%%%%%%%%%%%%%%%%%%%%%%%%%%%%%%%%%%%%%%%%%%%%%%%%%%%%%%%%%%
\begin{figure*}[htb]
	\centering
	\includegraphics[width=\linewidth]{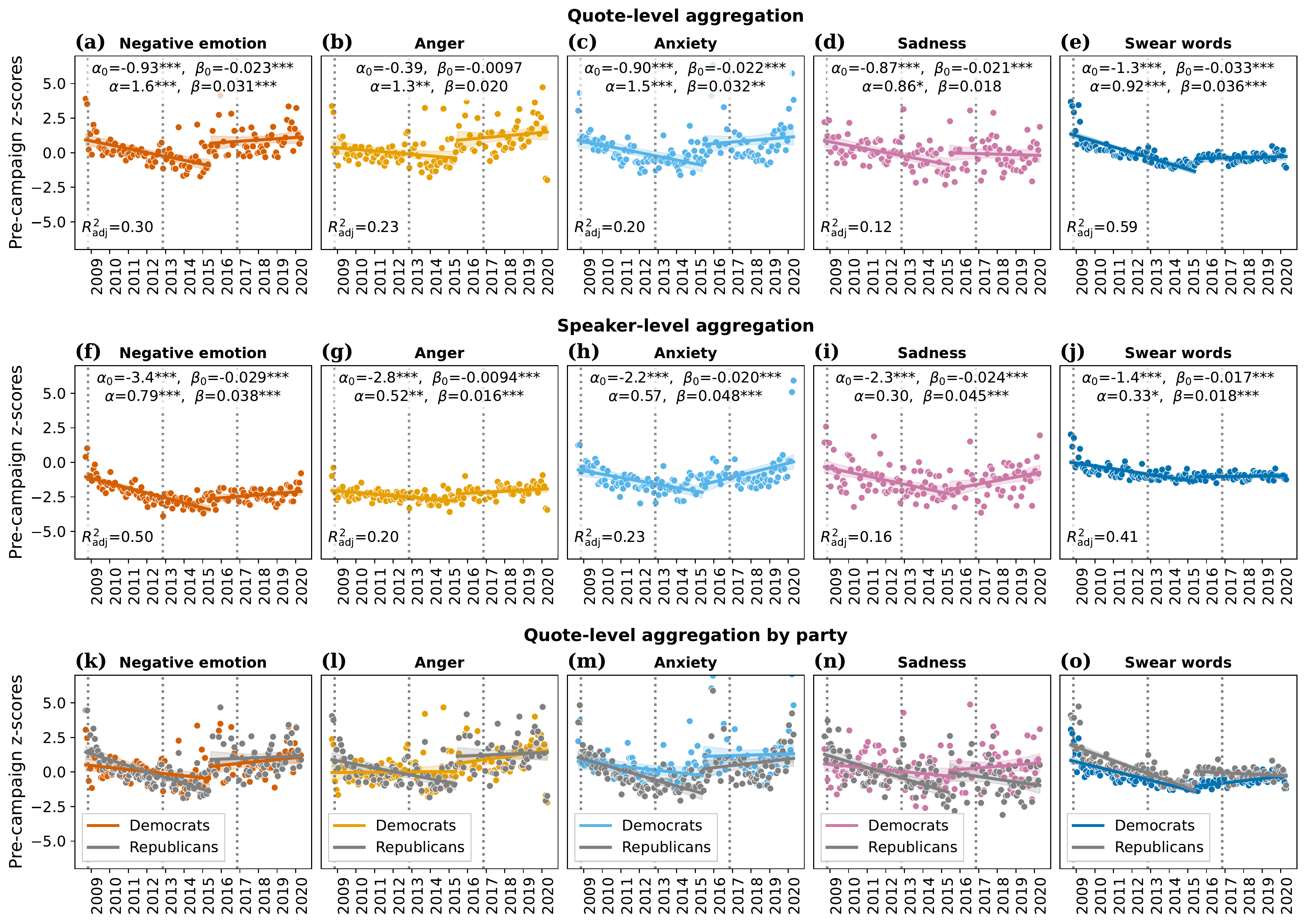}
	\caption{
	\textbf{Temporal evolution of negative language.}
	Columns correspond to negative-language word categories from LIWC;
	rows correspond to aggregation methods for computing monthly averages.
	Points show monthly averages, expressed as pre\hyp campaign $z$-scores (\ie, subtracting pre\hyp campaign mean from raw frequency values, and dividing by pre\hyp campaign standard deviation).
	Lines (with 95\% confidence intervals) were obtained via ordinary least squares regression, with coefficients shown in legends
	(\cf\ \Eqnref{eqn:regression} and \Figref{fig:fig0}(b) for interpretation of coefficients; tabular summary in \SupTabs{} S4, S6, S8, S9).
	\textbf{\mbox{(a--e)}}~\textit{Quote-level aggregation} micro\hyp averages over all quotes per month, \ie, speakers have weight proportional to their number of quotes in the respective month. Panel (a) shows the same data as \Figref{fig:fig0}.
	\textbf{(f--j)}~\textit{Speaker-level aggregation} macro\hyp averages by speaker, \ie, all speakers with at least one quote in a given month have equal weight in that month.
	\textbf{(k--o)}~\textit{Quote-level aggregation by party} performs the analysis of (a--e), but separately for quotes from Democrats \vs\ Republicans (coefficients omitted for clarity; \cf\ \SupTabs{} S8 and S9).
	Significance of regression coefficients: ***~$p<0.001$, **~$p<0.01$, *~$p<0.05$.
	We observe drastic shifts toward a more negative tone at the modeled June 2015 discontinuity (Trump's campaign start).
	}
	\label{fig:emotions}
\end{figure*}
%%%%%%%%%%%%%%%%%%%%%%%%%%%%%%%%%%%%%%%%%%%%%%%%%%%%%%%%%%%%

\xhdr{Temporal evolution of negative language}
\Figref{fig:emotions} visualizes the evolution of negative language between September 2008 and April 2020, with
one row per aggregation method,
one column per word category,
and one point per monthly aggregate score.
In order to quantify the shape of the curves, we fitted ordinary least squares linear regression models with a discontinuity in June 2015, the starting month of Donald Trump's primary campaign.
For a given word category, we model the aggregate score $y_t$ for month $t$ as
\begin{align}
	y_t &= \alpha_0
	+ \beta_0 \,t
	+ \alpha \,i_{t}
	+ \beta \,i_{t} \,t
	+ \varepsilon_{t},
\label{eqn:regression}
\end{align}
where $t \in \{-81, \dots, 58\}$, with $t=0$ corresponding to June 2015;
$i_t$ indicates whether $t$ is located before \vs\ after the campaign start (\ie, $i_t=0$ for $t < 0$, and $i_t=1$ for $t \geq 0$);
and $\varepsilon_t$ is the residual error.
As illustrated in \Figref{fig:fig0}(b), the coefficient $\alpha$ captures the immediate jump coinciding with the campaign start,
and $\beta$, the change in slope,
such that $\beta_0$ and $\beta_0 + \beta$ describe the slopes of the regression lines before and after the campaign start, respectively.
\Figref{fig:emotions} plots the fitted regression lines, alongside 95\% confidence intervals.

We first focus on the time series of quote-level aggregates (\Figref{fig:emotions}(a--e)).
The regression coefficients indicate a significant, sudden increase in the relative frequency of negative emotion words (\Figref{fig:emotions}(a)) in June 2015, by $\alpha=1.6$ ($p=4.1\times 10^{-6}$) pre-campaign standard deviations (SD),
translating to a relative increase of 8\% over the pre-campaign mean (\cf\ \Tabref{tab:scores}).
All three subcategories of negative emotion words, as well as swear words, also saw significant jumps in June 2015:
anger, by 1.3 SD ($+11\%$ of the pre-campaign mean, $p=0.0015$, \Figref{fig:emotions}(b));
anxiety, by 1.5 SD ($+11\%$, $p=5.6\times 10^{-4}$, \Figref{fig:emotions}(c));
sadness, by 0.86 SD ($+4\%$, $p=0.021$, \Figref{fig:emotions}(d));
and swear words, by 0.92 SD ($+31\%$, $p=2.9\times 10^{-6}$, \Figref{fig:emotions}(e)).%
\footnote{Note that the pre-campaign regression line for swear words (\Figref{fig:emotions}(e)) underestimates the values just before the discontinuity, mostly due to outliers at the left boundary (2008/09). For swear words, the actual June 2015 jump should thus be considered to be smaller than the estimate $\alpha$.
}
The June 2015 discontinuity was not only associated with a sudden increase in negative language, but also with a change in slope:
whereas the frequency of negative emotion words (\Figref{fig:emotions}(a)) had steadily and significantly decreased over the first 6.5 years of Obama's tenure by
$\beta_0=-0.023$ SD per month ($p=9.5 \times 10^{-7}$),
\ie, by about a quarter SD per year,
this trend came to a halt in June 2015, with a (non-significantly) positive slope of $\beta_0+\beta=0.0076$ (compound $p=0.21$) from June 2015 onward.%
% b0 = -0.02; b = 0.03; varb0 = 0.0000199547; varb = 0.0000764766
% p = 2*(1-pnorm((b0+b)/sqrt(varb-varb0))) = 0.2082096
% Sanity check: 2*(1-pnorm(1.959964)) = 0.05
\footnote{%
These results hold when removing outlier months whose quote-level aggregate score lies more than three standard deviations from the mean (\SupFig{} S1, and \SupTabs{} S5 and S7).}

% The regression model for negative emotion words captures 30\% of the total variance (adjusted $R^2=0.30, F=19.59, p=1.6 \times 10^{-10}$), the model for swear words, 59\% (adjusted $R^2=0.59, F=63.8, p=1.1 \times 10^{-16}$). The model fits are less tight, but remain significant, for the three subcategories of negative emotion words (adjusted $R^2$ reported in \Figref{fig:emotions}).
% For all five word categories, modeling a June 2015 discontinuity (\Eqnref{eqn:regression}) results in vastly better fits than a simpler model, $y_t = \alpha_0 + \beta_0 \,t + \varepsilon_{t}$, with no discontinuity ($F=25.6, p=4.5 \times 10^{-10}$ for negative emotion words; $F>6.73, p<0.0017$ for the subcategories; $F=55.6, p=1.1 \times 10^{-11}$ for swear words).
% F / p scores comparing to LINEAR REGRESSION:
% Anxiety: (15.204695934655042, 1.1897583107600695e-06)
% Anger: (9.977164491989667, 9.403006808272885e-05)
% Sadness: (6.730143075030358, 0.0016627539698617078)
We emphasize that the June 2015 discontinuity was chosen \textit{ex ante} based on incoming hypotheses grounded in the general public's subjective impression that Donald Trump's entering the political scene had changed the tone of US politics~\autocite{pew2019discourse}.
A data-driven analysis, conducted \textit{ex post,} revealed that June 2015 is in fact the optimal discontinuity for modeling the data:
out of 140 regression models analogous to \Eqnref{eqn:regression}, but each using another one of the 140 months of our analysis period as the discontinuity, the model with the June 2015 discontinuity yielded the best fit for negative emotion words and anger (\SupFig{} S5).
(For anxiety and sadness, slightly better fits were obtained when using different discontinuities; for swear words, a 2009 discontinuity led to a better fit due to outliers around that time; see \SupFig{} S5.)

%%%%%%%%%%%%%%%%%%%%%%%%%%%%%%%%%%%%%%%%%%%%%%%%%%%%%%%%%%%%

\xhdr{Role of speaker prominence}
%
% While the above, quote-level analysis approximates the perception of US politicians' negative tone on behalf of a media consumer reading online news, that analysis implicitly gives more weight to more frequently quoted politicians in the monthly averages.
When repeating the analysis using speaker-level, rather than quote-level, aggregates, \ie, weighing all speakers equally when averaging,
all of the above effects persisted qualitatively, but were reduced quantitatively (\Figref{fig:emotions}(f--j)):
in each of the five word categories, the immediate increase $\alpha$ at the June 2015 discontinuity dropped by between one-third and one-half,
% from 1.6 to 0.8 (-50%) / 1.3 to 0.5 (-62%) / 1.5 to 0.6 (-60%) / 2 x 0.9 to 0.3 (-67%)
remaining significant for negative emotions, anger, and swear words ($p=9.1\times 10^{-5},$ 0.0025, and 0.030, respectively),
but becoming non-significant for anxiety and sadness ($p=0.11$ and $0.43$, respectively).
The change of slope observed in the quote-level analysis also persisted in the speaker-level analysis.

\begin{figure*}[htb]
	\centering
	\includegraphics[width=\linewidth]{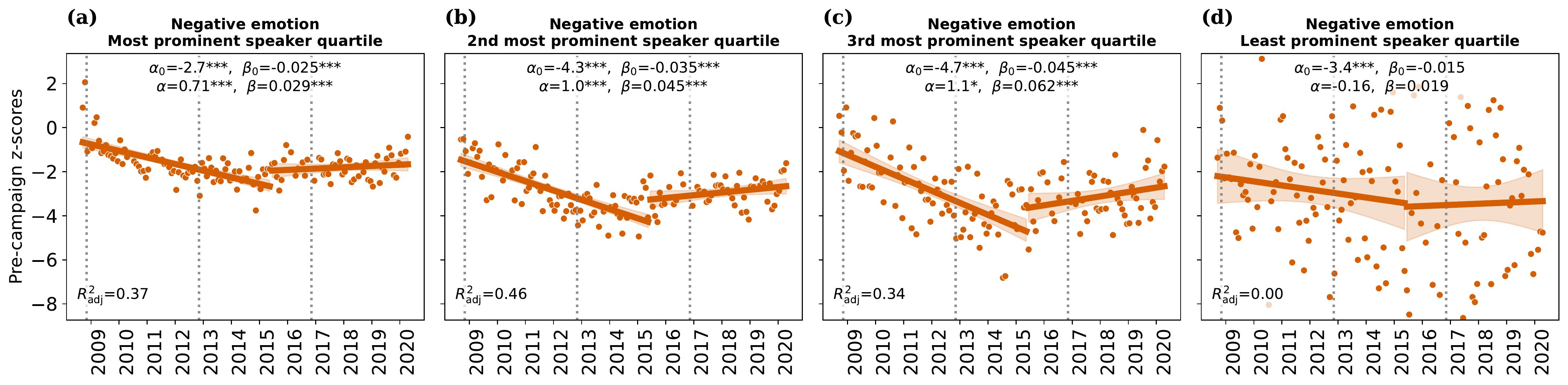}
	\caption{
		\textbf{Role of speaker prominence.}
		The set of 18,627 US politicians was split into four evenly-sized quartiles with respect to their total number of quotes (\ie, prominence);
		each panel shows the time series for negative emotion words obtained by performing monthly speaker-level aggregation on the respective quartile separately.
		That is, the figure shows the data of \Figref{fig:emotions}(f) after stratifying speakers by prominence.
		Lines (with 95\% confidence intervals) were obtained via ordinary least squares regression, with coefficients in legends (\cf\ \Eqnref{eqn:regression} and \Figref{fig:fig0}(b) for interpretation of coefficients).
		(Tabular summary of regression coefficients in \SupTabs{} S11, S12, S13, and S14.)
		Significance of regression coefficients: ***~$p<0.001$, **~$p<0.01$, *~$p<0.05$.
		We observe that the abrupt increase in negative emotion words emerges in all strata of speaker prominence except the least prominent stratum, and that quotes by more prominent speakers overall contain more negative emotion words.
		The figure focuses on one category of negative\hyp language words (negative emotion); for the other four categories, see \SupFig{} S3.
    }
	\label{fig:verbosity}
\end{figure*}

The fact that speaker-level effects are weaker than quote-level effects indicates that prominent, highly quoted speakers contribute disproportionally to the increase in negative language.
To confirm this conclusion more directly, we divided the speakers into four quartiles with respect to the number of quotes attributed to them, and repeated the speaker-level analysis for each quartile individually.
\Figref{fig:verbosity} shows
(1)~that the abrupt increase in negative emotion words emerges in all strata of speaker prominence except the least prominent stratum, and
(2)~that quotes by more prominent speakers generally contain more negative emotion words.
(Results for other word categories in \SupFig{} S3.)

%%%%%%%%%%%%%%%%%%%%%%%%%%%%%%%%%%%%%%%%%%%%%%%%%%%%%%%%%%%%
%%%%%%%%%%%%%%%%%%%%%%%%%%%%%%%%%%%%%%%%%%%%%%%%%%%%%%%%%%%%

%%%%%%%%%%%%%%%%%%%%%%%%%%%%%%%%%%%%%%%%%%%%%%%%%%%%%%%%%%%%

\xhdr{Biographic correlates of negative language}
The patterns identified above---a sudden increase in negative language in June 2015 followed by a change in slope---hold across party lines,%
\footnote{All $\alpha$ coefficients of \Figref{fig:emotions}(k--o) are positive. For Republicans, all are significant ($p<0.05$); for Democrats, the coefficients for anger ($\alpha=0.57$) and sadness ($\alpha=0.43$) are non-significant ($p>0.05$).}
but are more pronounced for Republicans,
as seen in \Figref{fig:emotions}(k--o), which tracks the (quote-level) evolution of negative language over time, analogously to \Figref{fig:emotions}(a--e), but separately for quotes by Democrats \vs\ Republicans.
For instance, the party-wise estimates of the June 2015 increase in negative emotion words is
$\alpha=0.89$ pre-campaign SD for Democrats ($p=0.014$)
and $\alpha=2.3$ SD for Republicans ($p=5.6 \times 10^{-9}$).

We further considered the possibility that the distribution of speaker characteristics may have changed over time; \eg, members of one party or gender may have become more frequently quoted over time.
To account for potential confounding due to such factors, we repeated the
% speaker-level
regression analysis with added control terms for four biographic attributes:
party affiliation (Republican, Democrat),
the party's federal role (Opposition, Government),
Congress membership (Non-Congress, Congress),
and gender (Male, Female).%
\footnote{Due to small sample size, we discarded speakers of a non-binary gender according to Wikidata.}
%\footnote{We counted all politicians with a US Congress Bio ID \autocite{congress.gov} as members of Congress, making no distinction between former and active members.}
For a given month, party affiliation fully determines the party's federal role, so for each month, the set of speakers can be partitioned into $2^3=8$ speaker groups, one per valid attribute combination.
We computed monthly aggregate scores $y_{gt}$ separately for each speaker group $g$, obtaining eight aggregate data points per month, and modeled them jointly using the following extended version of \Eqnref{eqn:regression}:
\begin{align}
	y_{gt} &= \alpha_0
	+ \beta_0 \,t
	+ \alpha \,i_{t}
	+ \beta \,i_{t} \,t
	+ \gamma \,\text{Democrat}_g
	+ \delta \,\text{Government}_g
	+ \zeta \,\text{Congress}_g
	+ \eta \,\text{Female}_g
	+ \varepsilon_{gt},
\label{eqn:extended_regression}
\end{align}
where $\text{Democrat}_g=1$ (or 0) if group $g$ contains Democrat (or Republican) speakers, and analogously for $\text{Government}_g$, $\text{Congress}_g$, and $\text{Female}_g$.

\begin{figure*}[htb]
	\centering
	\includegraphics[width=\linewidth]{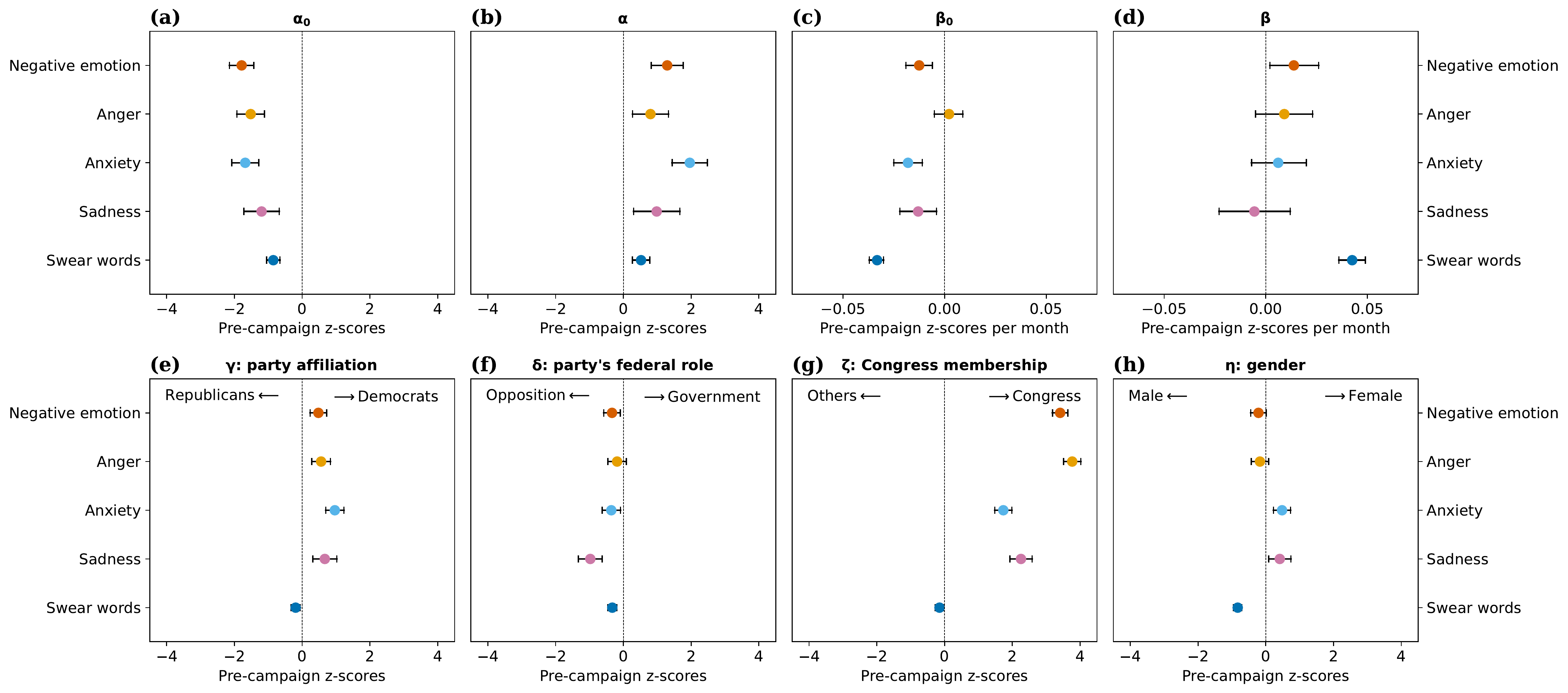}
	\caption{
	\textbf{Biographic correlates of negative language.}
	Coefficients (with 95\% confidence intervals) of ordinary least squares regression (\Eqnref{eqn:extended_regression}) for modeling time series of word categories (quote-level aggregation; speaker-level aggregation in \SupFig{} S6) while adjusting for 
	party affiliation ($\gamma$),
    the party's federal role ($\delta$),
    Congress membership ($\zeta$),
    and gender ($\eta$)
    (tabular summary in \SupTabs{} S15 and S16).
    Positive coefficients mark word categories that are, \textit{ceteris paribus,} used more frequently
    by Democrats than by Republicans,
    by members of the governing than by members of the opposition party,
    by Congress members than by others,
    or by females than by males
    (and \textit{vice versa} for negative coefficients).
    We observe that quotes by members of the opposition party, Congress members, and Democrats contain significantly more negative language.
    Importantly, the sudden June 2015 jump in negative language ($\alpha$) remains significant after adjusting for biographic attributes.
	}
	\label{fig:coefficients}
\end{figure*}

Inspecting the fitted coefficients (shown in \Figref{fig:coefficients} for quote-level aggregation; for speaker-level aggregation, see \SupFig{} S6), we make two observations:
First, the sudden June 2015 jump $\alpha$ in negative language (\Figref{fig:coefficients}(b)) appears even after adjusting for the four biographic attributes.
Second, we observe systematic correlations of negative language with biographic attributes, whereby quotes by members of the opposition party, Congress members, and Democrats contain, \textit{ceteris paribus,} significantly more negative language.
Moreover, quotes by females contained more anxiety and sadness, and quotes by males, more negative emotion, anger, and swear words.
The fact that quotes by members of the opposition party contain more negative language may reflect their role as corrective agents in the democratic process.
The fact that quotes by Congress members contain more negative language echoes previous work that has highlighted high-ranked politicians' role not only as deputies of their own party, but also as antagonists of the opposing party, epitomized in the finding that ``ideological moderates won't run'' \autocite{thomsen2014moderates} for Congress.
% Finally, the fact that quotes by Democrats contain more negative language appears to be at odds with an earlier finding that ``liberals display greater happiness'' \autocite{wojcik2015hapiness}.

%%%%%%%%%%%%%%%%%%%%%%%%%%%%%%%%%%%%%%%%%%%%%%%%%%%%%%%%%%%%
%%%%%%%%%%%%%%%%%%%%%%%%%%%%%%%%%%%%%%%%%%%%%%%%%%%%%%%%%%%%

%%%%%%%%%%%%%%%%%%%%%%%%%%%%%%%%%%%%%%%%%%%%%%%%%%%%%%%%%%%%

% when including all, not only signif, individuals:
% - Negemo: 117 vs. 83
% - Anger: 106 vs. 94
% - Sadness: 100 vs. 100
% - Anxiety: 131 vs. 69
% For all word categories, a majority of these 200 speakers (\eg, for negative emotion words, 117 \vs\ 83, or 24 \vs\ 16 when including only those whose $\alpha$ is significant with $p<0.05$) have a positive $\alpha$, \ie, most speakers' frequency of negative language increased abruptly in June 2015.
%
% Top 47 speakers: 77\% have pos alpha
% Top 4: 100\%
% Top 10: 70\%
% Top 50: 74\%
% Top 100: 63\%
% Top 200: 59\%

\begin{figure*}[htb]
	\centering
 	\includegraphics[width=\linewidth]{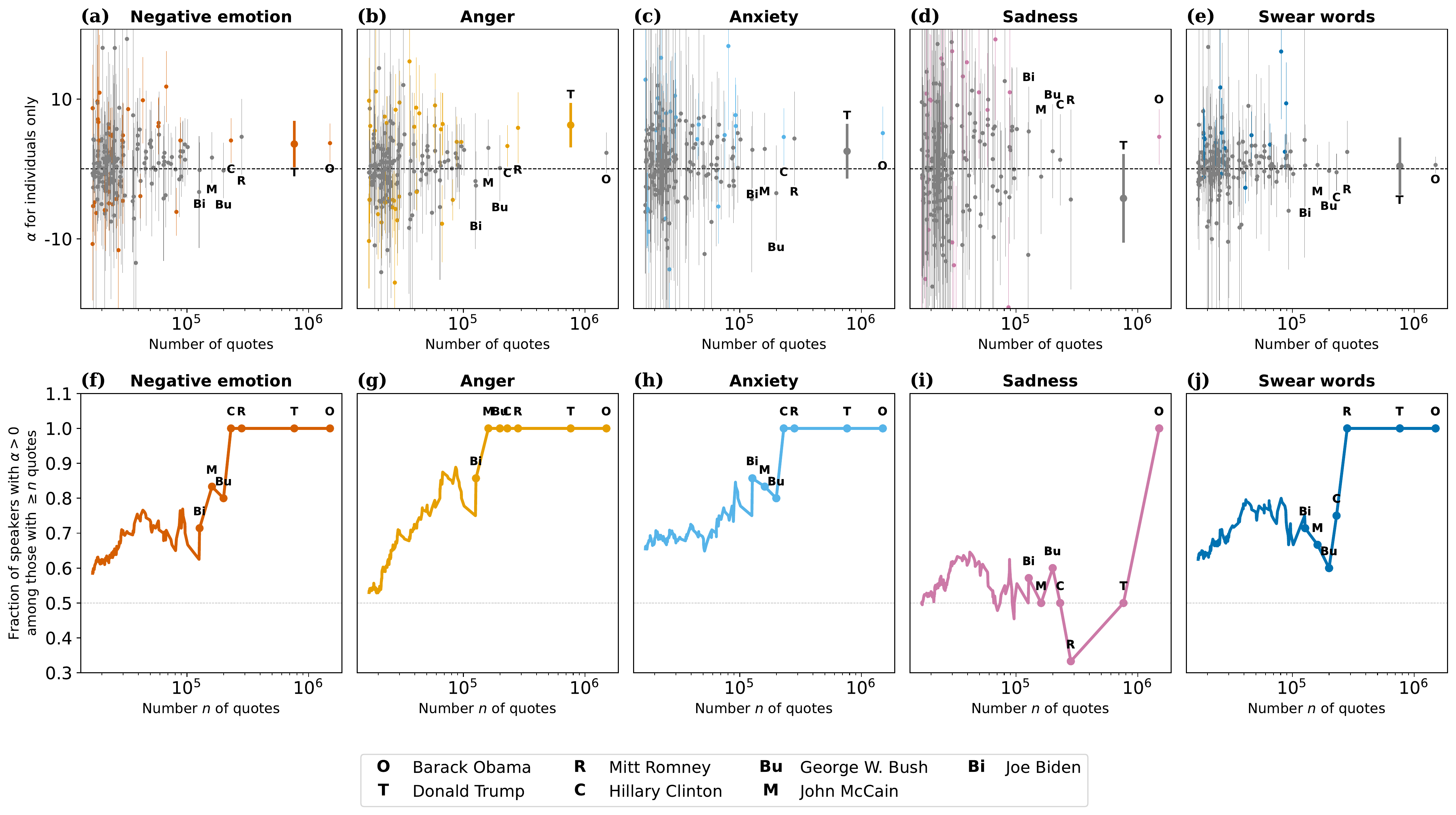}
	\caption{
	\textbf{Role of individual politicians: single-speaker study.}
	Results of ordinary least squares regressions (\Eqnref{eqn:regression}) fitted separately to the time series of each of the 200 most quoted speakers.
	\textbf{(a--e)}~Each speaker's $\alpha$ coefficient (capturing the size of the June 2015 jump, with 95\% confidence intervals) as a function of the speaker's number of quotes.
	Significant coefficients ($p<0.05$) in color, others in gray.
	\textbf{(f--j)}~Fraction of speakers with positive $\alpha$ among the speakers with at least $n$ quotes, as a function of $n$.
	We observe that, although many individual $\alpha$ coefficients are non-significant (a--e), the majority of coefficients are positive, particularly among the most-quoted speakers (as manifested in the increasing curves of (f--j)). That is, the June 2015 jump in negative language emerges even at the 
	individual level for a majority of the most-quoted politicians.
	}
	\label{fig:individuals_alpha}
\end{figure*}

\xhdr{Role of individual politicians}
In order to determine to what extent the above population-level effects mirror individual-level effects, we fitted regression models (\cf\ \Eqnref{eqn:regression}) to individual speakers' time series and analyzed the corresponding $\alpha$ coefficients (\ie, the size of the June 2015 jump; for completeness, $\beta$ in \SupFig{} S7).
In order to avoid data sparsity issues, this analysis focuses on the 200 most-quoted speakers, with $\alpha$ plotted in \Figref{fig:individuals_alpha}(a--e).
Additionally, \Figref{fig:individuals_alpha}(f--j) plots the fraction of speakers with
positive $\alpha$ among the speakers with at least $n$ quotes, as a function of $n$.
We observe that, although many individual $\alpha$ coefficients are non-significant ($p>0.05$, gray confidence intervals in \Figref{fig:individuals_alpha}(a--e)), the majority of coefficients are positive, particularly among the top most-quoted speakers (as manifested in the increasing curves of \Figref{fig:individuals_alpha}(f--j)).
For instance, for negative emotion words (\Figref{fig:individuals_alpha}(f)),
all of the top four most-quoted politicians have positive $\alpha$.
% (three of them with $p<0.05$).
Among the top 50, 74\% have positive $\alpha$;
among the top 100, 63\%;
and among the top 200, 59\%.
In other words, the June 2015 discontinuity emerges not only by aggregating at the population level, but mirrors a disruption that can also be perceived in the most-quoted politicians' individual language.
We further illustrate (\SupFig{} S8) using as examples the two presidents (Barack Obama, Donald Trump), vice presidents (Joe Biden, Mike Pence), and runners-up (Mitt Romney, Hillary Clinton) from the study period.
Four of these six speakers were associated with a significant ($p<0.05$) positive $\alpha$ for negative emotion words, and none with a significant negative $\alpha$.%
\footnote{Interestingly, although both Donald Trump ($\alpha=3.6$, $p=0.026$) and Barack Obama ($\alpha=3.7$, $p=0.0094$) followed the population-wide pattern by increasing their frequency of negative emotion words in June 2015, their pre- and post-discontinuity slopes were opposite to the population-wide pattern: their language first became more negative, then less negative.}
% \eg, Hillary Clinton's and Mike Pence's time series (\SupFig{} FIGREF) mirrors the overall population-level pattern,

\begin{figure*}[htb]
	\centering
	\includegraphics[width=\linewidth]{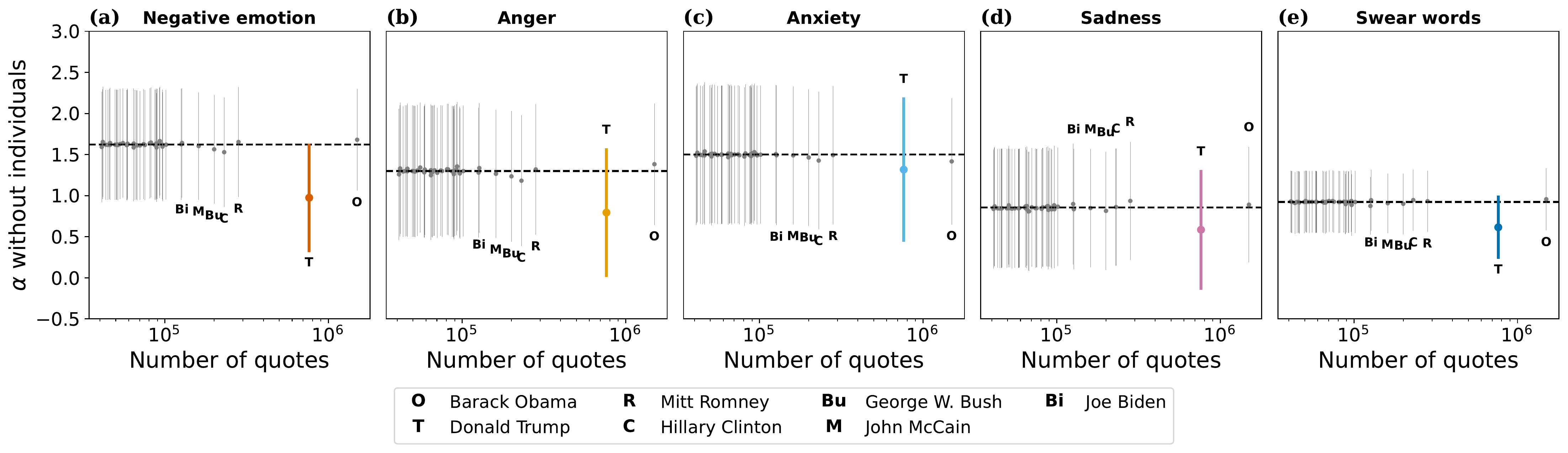}
	\caption{
	\textbf{Role of individual politicians: ablation study.}
	Results from ordinary least squares regression (\Eqnref{eqn:regression}; quote-level aggregation) on data sets obtained by removing all quotes by one target speaker and retaining all quotes from all 18,626 other speakers, using each of the 50 most quoted speakers as a target speaker.
	Each point shows the $\alpha$ coefficient (capturing the size of the June 2015 jump, with 95\% confidence intervals) obtained after removing the target speaker's quotes from the analysis, as a function of the target speaker's number of quotes.
	Dashed horizontal lines mark the coefficients obtained on the full data set without removing any speaker (\cf\ \Figref{fig:emotions}(a--e)).
	We observe that no single speaker's removal leads to a notable change in $\alpha$, except Donald Trump: \eg, by removing Trump's quotes, $\alpha$ drops by 40\% for negative emotion words (a).
	}
	\label{fig:ablation}
\end{figure*}

Next, we sought evidence whether individual politicians contributed particularly strongly to the overall increase in negative language.
We proceeded in an ablation study: for each of the 50 most quoted speakers, we repeated the quote-level regression (\Figref{fig:emotions}(a--e)) on a dataset consisting of all quotes except those from the respective speaker.
If $\alpha$ is particularly low when removing a given speaker, that speaker contributed particularly strongly to the overall June 2015 increase in negative language.
\Figref{fig:ablation}(a) shows that no single speaker's removal leads to an important change in $\alpha$ for negative emotion words, with one exception: Donald Trump.
By removing Donald Trump's quotes, the June 2015 increase in negative emotion words drops by 40\%, from $\alpha=1.6$ to $\alpha=0.98$ pre-campaign SD.%
\footnote{
More precisely, $(1.622-0.974)/1.622 = 0.40$.
When considering quotes by Republicans only (\cf\ \Figref{fig:emotions}(k--o)), $\alpha$ drops by 43\% when removing Trump's quotes, from 2.3 to 1.3 pre-campaign SD (\SupTabs{} S9 and S10).}
Put differently, by adding Donald Trump's quotes, the June 2015 increase in negative emotion words is boosted by 63\%.
Note that this is not merely an artifact of Trump's being quoted especially frequently: Obama was quoted about twice as frequently as Trump over the course of the 12-year period, yet removing his quotes does not notably affect the June 2015 increase in negative emotion words.
Qualitatively similar results hold for the other word categories, in particular for swear words (\Figref{fig:ablation}(e); $\beta$ in \SupFig{} S9 for completeness).
Although the size of the June 2015 jump in negative emotion words decreased drastically when removing Trump's quotes, note that it remained highly significant ($p=0.0032$).
That is, Trump was the main, but not the sole, driver of the effect.

% Numbers for Trump in ablation study (coefs with 95% CIs):
% Negative Emotions, ablation study (without Donald Trump):
% Alpha  = 1.0(0.332, 1.671), p = 3.219688e-03
% Beta   = 0.03(0.012, 0.045), p = 8.830058e-04
% Alpha0 = -1(-1.349, -0.553), p = 6.020761e-06
% Beta0  = -0.0231(-0.032, -0.015), p = 3.544148e-07

%%%%%%%%%%%%%%%%%%%%%%%%%%%%%%%%%%%%%%%%%%%%%%%%%%%%%%%%%%%%

\begin{figure*}[htb]
	\centering
	\includegraphics[width=\linewidth]{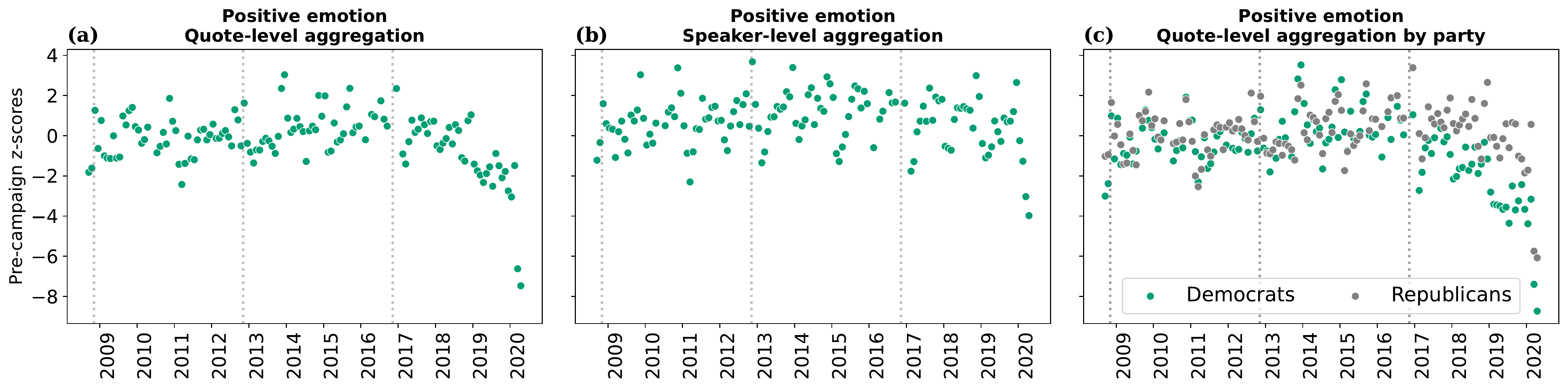}
	\caption{
	\textbf{Temporal evolution of positive emotion words.}
	Points show monthly averages, expressed as pre\hyp campaign $z$-scores.
	One panel per aggregation method for computing monthly averages.
    We observe that positive emotion words appear stable well into Trump's term, and then decline.
    % positive emotion words follow a pattern opposite to that of negative emotion words (\Figref{fig:emotions}(a, f, k)), with an initial rise followed by an eventual drop. Nothing particular is, however, observed for positive emotion words in June 2015. Rather, a turning point is located between mid 2018 and early 2019.
    }
	\label{fig:posemo}
\end{figure*}

\xhdr{Positive emotion words}
Complementary to the five word categories related to negative language analyzed above, we also conducted an exploratory analysis of positive language, as captured by LIWC's \textit{positive emotion} category.
As seen in \Figref{fig:posemo}(a), the time series of positive emotion words does not simply mirror that of negative emotion words. In particular, no particular changes are observed in June 2015. Rather, positive emotion words appear stable well into Trump's term, and then decline.
% The time series of positive emotion words (\Figref{fig:posemo}) exposes a pattern opposite to that of negative emotion words (\Figref{fig:emotions}(a, f, k)), with an initial rise followed by an eventual drop. As opposed to negative emotion words, however, nothing particular is observed for positive emotion words in June 2015. Rather, the rising trend appears to continue until Trump took office.
(The best regression fit is achieved when using July 2018 as the discontinuity, with a sharp drop in quality of fit after February 2019; \SupFig{} S10.)
As in the case of negative emotion words, the same pattern emerges within each party (\Figref{fig:posemo}(c)), with a more pronounced drop in positive emotion words for Democrats than for Republicans.
% As we had no incoming hypotheses about this time period, we relegate a deeper investigation to future work.

%%%%%%%%%%%%%%%%%%%%%%%%%%%%%%%%%%%%%%%%%%%%%%%%%%%%%%%%%%%%
%%%%%%%%%%%%%%%%%%%%%%%%%%%%%%%%%%%%%%%%%%%%%%%%%%%%%%%%%%%%
%%%%%%%%%%%%%%%%%%%%%%%%%%%%%%%%%%%%%%%%%%%%%%%%%%%%%%%%%%%%

\section*{Discussion}

The goal of this work has been to determine the accuracy of the average American's subjective impression (as of mid 2019)
that ``the tone and nature of political debate in the United States has become more negative in recent years'', and
that Donald Trump ``has changed the tone and nature of political debate [\dots{}] for the worse''~\autocite{pew2019discourse}.
Based on an analysis of 24 million quotes uttered by 18,627 US politicians between 2008 and 2020, we conclude that both of the above impressions are largely correct.

US politicians' tone indeed became suddenly and significantly more negative with the start of the 2016 primary campaigns in June 2015, and the frequency of negative language has remained elevated ever since.
Intriguingly, the shift at this incision point coincides with a similar abrupt shift in political polarization on online platforms \autocite{anderson2021social, frimer2022incivility}.
The sudden increase in negative language reported here was not only significant, but also strong; \eg, the frequency of negative emotion words jumped up by 1.6 pre-campaign standard deviations, or by 8\% of the pre-campaign mean.
The disruption becomes particularly stark when contrasted with the first 6.5 years of Obama's tenure, during which negative language had decreased steadily---at odds with a commonly held belief that Trump merely continued an older trend \autocite{fritze2021trump,james2021effects}.
The potential of negativity, incivility, and fear as tools to support political campaigns
% by increasing voter mobilization and media attention
has been long known \autocite{brooks2007incivility, gerstle2019negativity} and might explain the increase of negative language during the campaigns.
It cannot, however, explain the fact that the boost in negative language has continued for years after the campaigns had ended.

Rather, our results show that political debate during Donald Trump's entire term was characterized by a negative tone, and they specifically point to Trump as a key driver of this development:
when removing Trump's quotes from the corpus, the magnitude of the June 2015 jump in the frequency of negative emotion words drops by 40\%.
Interestingly, Trump's own negative tone (\SupFig{} S8(z)) followed long-term trends opposite to the population-wide trends, with an initial increase and a subsequent decrease (and with a sudden June 2015 increase akin to that of the population).
But as his language was overall far more negative than the average
(Trump's mean monthly average was
% 8.0107 (Mean over all months considered for the RDD)
8.0 pre-campaign standard deviations larger than the overall mean monthly average), he strongly skewed the overall tone toward the negative end when he moved to the center of the media's attention \autocite{patterson2017news}.

Despite Trump's disproportionate impact, the increase in negative language was, however, not due to Trump alone.
It remained significant in various complementary analyses:
when removing Trump's quotes from the analysis,
when giving equal weight to all speakers,
when analyzing each party separately,
and when analyzing the most-quoted speakers individually.
The negative tone of others might be partly provoked by Trump's statements and actions, but as we do not have access to a counterfactual world without Trump, our analysis cannot speak to this possibility.

Our analysis also cannot disambiguate to what extent the observed shift in negative language is caused by a real shift in what politicians say \vs\ a shift in what the media choose to report.
It is well known that media outlets are biased in what they report \autocite{gentzkow2010drives, puglisi2011newspaper, schiffer2006assessing}, typically towards negative news that tend to increase engagement on readers' behalf \autocite{grabe2006negativenews}, and this bias may have drifted during the 12 years analyzed here.
Future work may investigate this possibility by comparing quotes reported in the media to complete records of certain politicians' utterances in certain contexts, \eg, via Congressional records or public speeches \autocite{niculae2015quotus}.
(But whether the shift was caused by politicians or by the media, the effect would be identical: a more negative tone as perceived by news-reading citizens.)

Moreover, although word-counting is a powerful tool for detecting broad trends in large textual data \autocite{pennebaker2001personality, kahn2007emotions}, and although our results are robust to the specific choice of dictionary (Empath \autocite{fast2016empath} yields qualitatively identical results to LIWC \autocite{pennebaker2007liwc}; \SupFig{} S1, S2, S3, S4, and S11, and \SupTab{} S1), word-counting is crude and insensitive to nuances in context \autocite{beasley2015emotions}.
Our findings should thus be considered a starting point and hypothesis generator for more detailed analyses based on a closer inspection of the text and context of quotes, \eg, using a combination of advanced natural language processing tools and human annotation in order to shed light on the relation between negative language and polarization and partisanship: Has politicians' tone become more negative specifically when they talk about opponents?

The present study is also limited with respect to its time frame.
Although Quotebank \autocite{vaucher2021quotebank} is the largest existing corpus of speaker-attributed quotes in terms of size and temporal extent, the 12 years it spans are but a short sliver of the United States' long political history.
It is thus unclear when the decrease in negative language at the beginning of the corpus (during the initial 6.5 years of Obama's tenure) started.
Gentzkow \etal\ \autocite{gentzkow2019partisanship} observed a concurrent decrease in partisan language in Congressional speeches during Obama's tenure (2009--2016), which might be yet another manifestation of the processes that underlie the initial decrease in negative language observed here.
Gentzkow \etal's corpus of Congressional speeches, however, spans a much longer time (144 years, 1873--2016), during which overall partisanship increased much more---especially from the 1980s onward---than it eventually decreased during Obama's tenure.
We must therefore consider the possibility that, analogously, the 2009--2015 decrease in negative language may have been merely a short anomaly in a longer increasing trend in negative language.
(Unfortunately, Gentzkow \etal's corpus ends with Obama's tenure, so we cannot compare trends in negative language to trends in partisan language during Trump's tenure.)

We saw that the June 2015 rise in negative language was not accompanied by a simultaneous drop in positive language. Rather, maybe in line with a general positivity bias in human language \autocite{Dodds2015human}, positive language remained stable until it eventually dropped during Trump's term.
% Around the same time, anxiety-related words became more frequent (\Figref{fig:emotions}(c)).
What happened at this point is an open question that lies beyond the scope of this work.
The phenomenon highlights, however, that positive and negative emotion words are not necessarily complementary.
From the start of the 2016 primary campaigns through the first half of Donald Trump's term, political tone was both highly positive and highly negative---akin to Trump's own style, characterized by typical features of affective polarization such as positive self-representation and negative other-presentation \autocite{ahmadian2017trump, kreis2017tweets, lewandowsky2020using}:
% , exemplified by this 2013 tweet:
``Sorry losers and haters, but my I.Q.\ is one of the highest -and you all know it! Please don't feel so stupid or insecure,it's not your fault''.%
\footnote{\url{https://web.archive.org/web/20151013150633/https://twitter.com/realDonaldTrump/status/332308211321425920}}

This said, a key contribution of this work is the conclusion that, despite Donald Trump's key role in setting the tone of political debate, the shift towards a more negative tone permeates all of US politics.
The consequences are tangible, as shown by research that highlights the detrimental effects of affective polarization on altruism \autocite{whitt2020tribalism}, trust \autocite{mutz2005videomalaise}, and opinion formation \autocite{druckman2021affect, forgette2006television},
and by polls showing that politics has become a stressful experience for Americans \autocite{pew2019discourse, apa2017coping}, exacting an ever increasing toll on their physical, emotional, and social well-being \autocite{Smith2019cost, apa2018genZ}.
Finding ways to break out of this cycle of negativity is one of the big challenges faced by the United States today.

%%%%%%%%%%%%%%%%%%%%%%%%%%%%%%%%%%%%%%%%%%%%%%%%%%%%%%%%%%%%
%%%%%%%%%%%%%%%%%%%%%%%%%%%%%%%%%%%%%%%%%%%%%%%%%%%%%%%%%%%%
%%%%%%%%%%%%%%%%%%%%%%%%%%%%%%%%%%%%%%%%%%%%%%%%%%%%%%%%%%%%

\section*{Materials and Methods}

\xhdr{US politicians}
We considered as politicians all people for whom the Wikidata knowledge base \autocite{wikidata} (version of 27 October 2021) lists ``politician'' (Wikidata item Q82955) or a subclass thereof as an occupation (P106).
Given our focus on the United States, we included only those politicians whose party affiliation (P102) was listed as Democrat (Q29552) or Republican (Q29468).
We considered as members of Congress those for whom Wikidata listed a US Congress Bio ID%
\footnote{For instance, \url{https://bioguide.congress.gov/search/bio/S000148}}
(P1157), making no distinction between former and active members of Congress.
Due to the small number of speakers of a non-binary gender, we included only speakers whose gender (P21) was listed in Wikidata as male (Q6581097) or female (Q6581072).

\xhdr{Quotebank}
The analyzed quotes were obtained from Quotebank \autocite{vaucher2021quotebank}, a publicly available \autocite{vaucher2021zenodo} corpus of 235 million unique speaker-attributed quotes extracted from 127 million English news articles published between September 2008 and April 2020, provided by the large-scale online media aggregation service Spinn3r.com.
While Spinn3r.com collects and supplies content from a comprehensive set of news domains \autocite{west2021postmortem}, it also includes much content beyond news alone, including ``social media, weblogs, news, video, and live web content'' \autocite{spinn3r_doc}.
Therefore, Quotebank was extracted from a filtered data set consisting only of content from a set of about 17,000 online news domains, defined as the set of domains appearing at least once in the large \textit{News on the Web} corpus \autocite{nowCorpus}, which has been collecting large numbers of news articles from Google News and Bing News since 2010 and may thus be considered to provide a comprehensive list of English-language media outlets.
We emphasize that the \textit{News on the Web} corpus was only used for defining the set of news domains. It was not used for obtaining the news articles themselves, which originated from Spinn3r.com only.

%%% When filtering Spinn3r, we kept only the 17,685 host-level domains that are also indexed by the NOW corpus.

We use the quote-centric (as opposed to the article-centric) version of Quotebank, which contains one entry per unique quote and aggregates information from all news articles in which the quote occurs.
% We refer to the Quotebank paper \autocite{vaucher2021quotebank} and data repository \autocite{vaucher2021zenodo} for full details about the corpus and restrict ourselves to summarizing the most relevant facts here.
In constructing Quotebank, a machine learning algorithm (based on the large pre-trained BERT language model \autocite{DevlinCLT19bert}) was used to infer, for each quote, a probability distribution over all speaker names mentioned in the text surrounding the quote (and an additional ``no speaker'' option), specifying each speaker's estimated probability of having uttered the quote.
For a given quote, we maintained only the name with the highest probability and consider it to indicate the speaker of the quote (a heuristic that was shown to have an accuracy of around 87\% \autocite{vaucher2021quotebank}).
A speaker name may be ambiguous. In such cases, Quotebank does not attempt to disambiguate the name,
but rather provides a list of all speakers to whom the name may refer, where speakers are identified by their unique ID from the Wikidata knowledge base \autocite{wikidata}.
In our analysis, we attributed quotes that were linked to ambiguous speaker names (less than 6\% of all quotes, see below) to each speaker to whom the respective name may refer.

To further clean the data set, we discarded quotes that were clearly non-verbal (\eg, consisting of URLs, HTML tags, or dates only).
Moreover, on some days, Spinn3r.com, which provided the news articles for Quotebank, failed to deliver content due to technical problems \autocite{vaucher2021quotebank}.
We therefore identified missing days as those having less than 10\% of the median number of unique quotes and dropped eight (out of 140) months with 20 or more missing days:
May 2010,
June 2010,
January 2016,
March 2016,
June 2016,
October 2016,
November 2016,
January 2017.
% '2010-5', '2010-6', '2016-1', '2016-3', '2016-6', '2016-10', '2016-11', '2017-1'
% This equals removing all months with less than 50,000 monthly political quotes. (\SubFig{FIGREF})

\xhdr{Quotes by US politicians}
Keeping only quotes attributed to US politicians (see above), we obtained 24 million quotes attributed to 18,627 unique US politicians.
% (out of a total of 1,147,898 speakers appearing in all of Quotebank).
Out of these,
4,487 were female, 14,140 male;
9,390 were Democrats, 9,237 Republicans;%
\footnote{
Out of an original set of 18,954 US politicians appearing in Quotebank, Wikidata listed 327 as (former) members of both parties, usually because they switched membership during their careers. Out of these 327, we manually checked the 21 politicians with over 10,000 quotes, of whom 16 could be unambiguously assigned to one party for the study period. The remaining 311 politicians were dropped from the analysis.}
and 1,790 were labeled as members of Congress.
% 0	Q76	140	1499080	Barack Obama
% 1	Q22686	138	762777	Donald Trump
% 2	Q4496	125	281413	Mitt Romney
% 3	Q6294	137	230212	Hillary Clinton
% 4	Q207	133	199909	George W. Bush
% 5	Q10390	133	160688	John McCain
% 6	Q6279	133	127004	Joe Biden
The most prolific speakers were
Barack Obama (1.5m quotes),%
\footnote{It seems unreasonable that any single person could utter over 300 different quote-worthy statements a day. Instead, the large number can be explained by news outlets attending to different parts of a politician's spoken output. For instance, a long sentence can be quoted in various ways.}
Donald Trump (763k quotes),
Mitt Romney (281k quotes),
Hillary Clinton (230k quotes),
George W.\ Bush (200k quotes),
John McCain (161k quotes),
and Joe Biden (127k quotes).
For a list of the 30 most frequently quoted politicians, see \SupTab{} S17.
Although, as mentioned, ambiguous names led to some quotes being attributed to multiple speakers, this happened rarely:
% Quotes attributed to n people (out of 23,968,098 quotes) [total attributions - 25,691,525]:
% 1: 22595960, 94.3%
% 2: 1052446, 4.4%
% 3: 288095, 1.2%
% 4: 31597, 0.13%
the vast majority (94.3\%) of quotes were attributed to a single politician,
4.3\% to two politicians,
1.2\% to three politicians,
and 0.13\% to four politicians.
\SupFig{} S12 shows the number of quotes and the number of unique speakers per month.

\xhdr{Aggregation methods}
Consider a fixed LIWC word category $c$ and a fixed month $t$.
Let $S$ be the set of speakers with at least one quote during month $t$.
Let $Q_s$ be the set of quotes attributed to speaker $s \in S$ during month $t$,
and let $Q = \bigcup_{s \in S} Q_s$ be the set of all quotes from month $t$.%
\footnote{The (few) quotes attributed to multiple speakers are included in $Q$ once per speaker.}
Let $\score{q}$ be quote $q$'s score for word category $c$.

Then, the \textit{quote-level aggregate score} for word category $c$ in month $t$ is defined as
% \begin{gather}
% \begin{aligned}
$$ y^\text{quote}_t
= \frac{1}{|Q|} \sum_{q \in Q} \score{q}
= \sum_{s \in S} \underbrace{\frac{|Q_s|}{|Q|}}_\text{depends on $s$}  \underbrace{\left[ \frac{1}{|Q_s|} \sum_{q \in Q_s} \score{q} \right]}_\text{average of speaker $s$},$$
% \end{aligned}
% \end{gather}
and the \textit{speaker-level aggregate score,} as
$$ y^\text{speaker}_t
= \frac{1}{|S|} \sum_{s \in S} \left( \frac{1}{|Q_s|} \sum_{q \in Q_s} \score{q} \right)
= \sum_{s \in S} \underbrace{\frac{1}{|S|}}_{\text{constant}} \underbrace{\left[ \frac{1}{|Q_s|} \sum_{q \in Q_s} \score{q} \right]}_\text{average of speaker $s$}. $$
That is, in quote-level aggregation, every speaker contributes with weight proportional to their number of quotes,
whereas in speaker-level aggregation, all speakers contribute with equal weight.

\xhdr{Data availability}
The Quotebank corpus is publicly available on Zenodo at \url{https://doi.org/10.5281/zenodo.4277311}.
Aggregated data derived from Quotebank are available on GitHub at \url{https://github.com/epfl-dlab/Negativity_in_2016_campaign}.

\xhdr{Code availability}
All analysis code is available on GitHub at \url{https://github.com/epfl-dlab/Negativity_in_2016_campaign}.

\printbibliography

% \renewcommand\refname{References}
% \bibliographystyle{plain}
% \bibliography{bibliography}

\clearpage
\appendix
\renewcommand\thefigure{A\arabic{figure}}
\setcounter{figure}{0}



\section*{Supplementary material}

Table 2: **The most common 30 words for each word category, before 15 June 2015**. Bold values deviate from the second period (after 15 June 2015) by more than one percentage point

| Top N words | Negative emotion | | Anger | | Anxiety | | Sadness | | Swear words | |
|---|---|---|---|---|---|---|---|---|---|---|
| 1 | problem.* | 4.35% | **fight.*** | **7.67%** | risk.* | 10.84% | low.* | 10.88% | **hell** | **17.14%** |
| 2 | numb.* | 3.20% | attack.* | 4.98% | terror.* | 8.35% | fail.* | 9.85% | **damn.*** | **8.68%** |
| 3 | fight.* | 2.67% | threat.* | 4.93% | worr.* | 5.89% | lost | 7.37% | heck | 6.72% |
| 4 | low.* | 2.11% | cut | 4.87% | struggl.* | 5.21% | hurt.* | 5.22% | screw.* | 6.62% |
| 5 | bad | 1.92% | **war** | **4.83%** | pressur.* | 5.16% | lose | 5.11% | dumb.* | 6.47% |
| 6 | fail.* | 1.91% | defens.* | 4.21% | doubt.* | 4.43% | loss.* | 4.23% | **dick** | **6.20%** |
| 7 | difficult.* | 1.87% | kill.* | 4.03% | avoid.* | 4.17% | disappoint.* | 4.12% | ass | 4.95% |
| 8 | tough.* | 1.79% | critical | 3.71% | fear | 3.52% | damag.* | 3.88% | **shit.*** | **4.86%** |
| 9 | wrong.* | 1.77% | danger.* | 3.10% | horr.* | 2.50% | alone | 2.89% | crap | 3.41% |
| 10 | attack.* | 1.73% | weapon.* | 2.65% | overwhelm.* | 2.47% | losing | 2.61% | fuck | 3.07% |
| 11 | risk.* | 1.73% | argu.* | 2.45% | vulnerab.* | 2.38% | overwhelm.* | 2.03% | fuckin.* | 2.92% |
| 12 | threat.* | 1.71% | victim.* | 2.21% | afraid | 2.34% | devastat.* | 2.01% | butt | 2.85% |
| 13 | cut | 1.69% | violat.* | 2.20% | craz.* | 2.34% | defeat.* | 1.96% | bitch.* | 2.75% |
| 14 | war | 1.68% | battl.* | 2.01% | uncertain.* | 2.18% | miss | 1.91% | darn | 2.68% |
| 15 | serious | 1.51% | offens.* | 1.99% | desperat.* | 1.97% | traged.* | 1.90% | bloody | 2.28% |
| 16 | defens.* | 1.46% | abuse.* | 1.87% | stress.* | 1.96% | reject.* | 1.89% | piss.* | 2.09% |
| 17 | lost | 1.43% | aggress.* | 1.68% | scare.* | 1.87% | sad | 1.81% | suck | 2.04% |
| 18 | kill.* | 1.40% | frustrat.* | 1.53% | confus.* | 1.76% | missed | 1.61% | sucks | 1.43% |
| 19 | terror.* | 1.33% | destroy.* | 1.42% | upset.* | 1.75% | suffering | 1.31% | sucked | 1.22% |
| 20 | critical | 1.29% | hate | 1.23% | shame.* | 1.71% | depress.* | 1.30% | bastard.* | 1.20% |
| 21 | danger.* | 1.08% | blam.* | 1.19% | distract.* | 1.62% | missing | 1.30% | butts | 1.02% |
| 22 | hurt.* | 1.01% | critici.* | 1.06% | embarrass.* | 1.61% | resign.* | 1.23% | goddam.* | 0.78% |
| 23 | lose | 0.99% | fought | 1.00% | guilt.* | 1.33% | regret.* | 1.23% | asshole.* | 0.78% |
| 24 | worr.* | 0.94% | murder.* | 1.00% | emotional | 1.27% | broke | 1.21% | motherf.* | 0.68% |
| 25 | weapon.* | 0.92% | assault.* | 0.98% | shake.* | 1.21% | tragic.* | 1.16% | nigger.* | 0.62% |
| 26 | mistak.* | 0.89% | outrag.* | 0.92% | disturb.* | 1.21% | abandon.* | 1.06% | queer.* | 0.55% |
| 27 | argu.* | 0.85% | violent.* | 0.88% | anxi.* | 1.13% | isolat.* | 1.00% | boob.* | 0.49% |
| 28 | struggl.* | 0.83% | hell | 0.87% | alarm.* | 1.04% | empt.* | 0.95% | fucked.* | 0.48% |
| 29 | pressur.* | 0.82% | punish.* | 0.85% | nervous.* | 1.02% | suffer | 0.93% | dang | 0.47% |
| 30 | loss.* | 0.82% | offend.* | 0.83% | suspicio.* | 0.92% | grave.* | 0.93% | prick.* | 0.47% |
| Total nr Expressions | 499 | | 184 | | 91 | | 101 | | 53 | |
| Total Matches | 5763501 | | 2005132 | | 918339 | | 1117453 | | 101773 | |

Table 3: **The most common 30 words for each word category, after 15 June 2015**. Bold values deviate from the first period (before 15 June 2015) by more than one percentage point

| Top N words | Negative emotion | | Anger | | Anxiety | | Sadness | | Swear words | |
|---|---|---|---|---|---|---|---|---|---|---|
| 1 | problem.* | 3.79% | **fight.*** | **9.15%** | risk.* | 10.82% | low.* | 11.05% | **hell** | **24.09%** |
| 2 | fight.* | 3.36% | threat.* | 5.72% | terror.* | 8.92% | fail.* | 9.05% | **damn.*** | **11.48%** |
| 3 | numb.* | 3.03% | attack.* | 5.43% | worr.* | 5.50% | lost | 6.89% | dumb.* | 7.11% |
| 4 | threat.* | 2.10% | critical | 3.90% | pressur.* | 4.94% | hurt.* | 5.38% | screw.* | 6.71% |
| 5 | low.* | 2.04% | **danger.*** | **3.80%** | horr.* | 4.27% | lose | 4.95% | heck | 6.08% |
| 6 | attack.* | 1.99% | kill.* | 3.60% | struggl.* | 4.26% | damag.* | 4.22% | **shit.*** | **4.71%** |
| 7 | bad | 1.95% | war | 3.59% | fear | 4.03% | loss.* | 3.78% | ass | 4.47% |
| 8 | risk.* | 1.78% | defens.* | 2.94% | doubt.* | 3.91% | disappoint.* | 3.73% | **crap** | **3.43%** |
| 9 | wrong.* | 1.78% | weapon.* | 2.82% | vulnerab.* | 3.42% | alone | 2.90% | darn | 2.75% |
| 10 | fail.* | 1.67% | cut | 2.79% | avoid.* | 3.20% | devastat.* | 2.57% | fuckin.* | 2.51% |
| 11 | terror.* | 1.47% | victim.* | 2.74% | craz.* | 2.54% | traged.* | 2.34% | bitch.* | 2.47% |
| 12 | critical | 1.43% | abuse.* | 2.41% | overwhelm.* | 2.38% | losing | 2.32% | piss.* | 2.42% |
| 13 | difficult.* | 1.43% | violat.* | 2.26% | afraid | 2.34% | defeat.* | 2.31% | butt | 2.37% |
| 14 | tough.* | 1.43% | hate | 1.94% | shame.* | 1.95% | sad | 2.19% | suck | 2.18% |
| 15 | danger.* | 1.40% | argu.* | 1.87% | shy.* | 1.87% | overwhelm.* | 2.12% | bloody | 2.17% |
| 16 | serious | 1.38% | offens.* | 1.86% | uncertain.* | 1.85% | reject.* | 2.00% | sucks | 1.70% |
| 17 | kill.* | 1.32% | destroy.* | 1.54% | desperat.* | 1.79% | resign.* | 1.66% | fuck | 1.62% |
| 18 | war | 1.32% | battl.* | 1.53% | stress.* | 1.77% | suffering | 1.60% | sucked | 1.39% |
| 19 | lost | 1.28% | aggress.* | 1.52% | confus.* | 1.65% | miss | 1.59% | motherf.* | 1.09% |
| 20 | defens.* | 1.08% | assault.* | 1.44% | embarrass.* | 1.56% | missed | 1.54% | queer.* | 0.95% |
| 21 | weapon.* | 1.04% | frustrat.* | 1.43% | disturb.* | 1.55% | tragic.* | 1.47% | butts | 0.80% |
| 22 | cut | 1.02% | murder.* | 1.27% | scare.* | 1.53% | missing | 1.29% | bastard.* | 0.78% |
| 23 | victim.* | 1.01% | fought | 1.08% | distract.* | 1.51% | regret.* | 1.15% | goddam.* | 0.73% |
| 24 | hurt.* | 1.00% | violent.* | 1.06% | guilt.* | 1.50% | isolat.* | 1.13% | pussy.* | 0.58% |
| 25 | lose | 0.92% | blam.* | 0.99% | upset.* | 1.48% | abandon.* | 1.11% | tit | 0.53% |
| 26 | worr.* | 0.91% | hell | 0.95% | alarm.* | 1.29% | broke | 1.04% | asshole.* | 0.52% |
| 27 | abuse.* | 0.89% | critici.* | 0.93% | emotional | 1.22% | grave.* | 1.01% | asses | 0.49% |
| 28 | violat.* | 0.83% | outrag.* | 0.92% | anxi.* | 1.17% | suffer | 0.92% | dick | 0.47% |
| 29 | mistak.* | 0.82% | harass.* | 0.92% | shake.* | 0.95% | suffered | 0.87% | dang | 0.44% |
| 30 | pressur.* | 0.81% | punish.* | 0.79% | tension.* | 0.80% | empt.* | 0.83% | crappy | 0.42% |
| Total nr Expressions | 499 | | 184 | | 91 | | 101 | | 53 | |
| Total Matches | 3695279 | | 1357629 | | 608580 | | 683988 | | 53284 | |

Table 4: **OLS regression parameters for quote-level aggregation shown in Figure 2 in the main text**. SEs of coefficients are in parantheses. ***$p < 0.001$, **$p < 0.01$ and *$p < 0.05$

| | negative emotion | | anger | | anxiety | | sadness | | swear words | |
|---|---|---|---|---|---|---|---|---|---|---|
| $\alpha_0$ | -0.932***(0.209) | | -0.391 | (0.249) | -0.903***(0.263) | | -0.867***(0.226) | | -1.347***(0.117) | |
| $\alpha$ | 1.622***(0.337) | | 1.299** | (0.402) | 1.501***(0.424) | | 0.855* | (0.365) | 0.923***(0.189) | |
| $\beta_0$ | -0.023***(0.004) | | -0.010 | (0.005) | -0.022***(0.006) | | -0.021***(0.005) | | -0.033***(0.002) | |
| $\beta$ | 0.031***(0.009) | | 0.020 | (0.010) | 0.032** (0.011) | | 0.018 | (0.009) | 0.036***(0.005) | |
| $R^2$ | 0.315 | | 0.251 | | 0.215 | | 0.136 | | 0.599 | |
| Adj. $R^2$ | 0.299 | | 0.234 | | 0.197 | | 0.116 | | 0.590 | |
| No. Observations | 132 | | 132 | | 132 | | 132 | | 132 | |

Table 5: **OLS regression parameters for quote-level aggregation, excluding outliers**. SEs of coefficients are in parantheses. ***$p < 0.001$, **$p < 0.01$ and *$p < 0.05$

|  | negative emotion | anger | anxiety | sadness | swear words |
|---|---|---|---|---|---|
| $\alpha_0$ | -0.780***(0.184) | -0.391 (0.231) | -0.816***(0.187) | -0.944***(0.209) | -1.110***(0.075) |
| $\alpha$ | 1.222***(0.299) | 1.260** (0.382) | 0.989** (0.317) | 0.756* (0.340) | 0.686***(0.120) |
| $\beta_0$ | -0.017***(0.004) | -0.010 (0.005) | -0.019***(0.004) | -0.022***(0.004) | -0.024***(0.002) |
| $\beta$ | 0.031***(0.008) | 0.016 (0.010) | 0.031***(0.008) | 0.023* (0.009) | 0.027***(0.003) |
| $R^2$ | 0.347 | 0.224 | 0.250 | 0.168 | 0.632 |
| Adj. $R^2$ | 0.331 | 0.206 | 0.232 | 0.149 | 0.623 |
| No. Observations | 129 | 129 | 127 | 130 | 128 |

Table 6: **OLS regression parameters for speaker-level aggregation shown in Figure 2 in the main text**. SEs of coefficients are in parantheses. ***$p < 0.001$, **$p < 0.01$ and *$p < 0.05$

|  | negative emotion | anger | anxiety | sadness | swear words |
|---|---|---|---|---|---|
| $\alpha_0$ | -3.413***(0.121) | -2.800***(0.105) | -2.170***(0.223) | -2.268***(0.238) | -1.374***(0.092) |
| $\alpha$ | 0.790***(0.195) | 0.522** (0.169) | 0.573 (0.359) | 0.304 (0.384) | 0.328* (0.149) |
| $\beta_0$ | -0.029***(0.003) | -0.009***(0.002) | -0.020***(0.005) | -0.024***(0.005) | -0.017***(0.002) |
| $\beta$ | 0.038***(0.005) | 0.016***(0.004) | 0.048***(0.009) | 0.045***(0.010) | 0.018***(0.004) |
| $R^2$ | 0.509 | 0.220 | 0.251 | 0.181 | 0.421 |
| Adj. $R^2$ | 0.498 | 0.202 | 0.234 | 0.161 | 0.408 |
| No. Observations | 132 | 132 | 132 | 132 | 132 |

Table 7: **OLS regression parameters for speaker-level aggregation, excluding outliers**. SEs of coefficients are in parantheses. ***$p < 0.001$, **$p < 0.01$ and *$p < 0.05$

|  | negative emotion | anger | anxiety | sadness | swear words |
|---|---|---|---|---|---|
| $\alpha_0$ | -3.360***(0.115) | -2.800***(0.105) | -2.170***(0.150) | -2.206***(0.219) | -1.374***(0.092) |
| $\alpha$ | 0.737***(0.184) | 0.522** (0.169) | 1.010***(0.245) | 0.346 (0.354) | 0.328* (0.149) |
| $\beta_0$ | -0.027***(0.002) | -0.009***(0.002) | -0.020***(0.003) | -0.021***(0.005) | -0.017***(0.002) |
| $\beta$ | 0.036***(0.005) | 0.016***(0.004) | 0.026***(0.007) | 0.036***(0.009) | 0.018***(0.004) |
| $R^2$ | 0.498 | 0.220 | 0.286 | 0.153 | 0.421 |
| Adj. $R^2$ | 0.486 | 0.202 | 0.269 | 0.132 | 0.408 |
| No. Observations | 131 | 132 | 130 | 129 | 132 |

Table 8: **OLS regression parameters for quote-level aggregation for Democrats shown in Figure 2 in the main text**. SEs of coefficients are in parantheses. ***$p < 0.001$, **$p < 0.01$ and *$p < 0.05$

|  | negative emotion | anger | anxiety | sadness | swear words |
|---|---|---|---|---|---|
| $\alpha_0$ | -0.481* (0.223) | 0.041 (0.269) | -0.247 (0.296) | -0.297 (0.289) | -1.397***(0.110) |
| $\alpha$ | 0.892* (0.360) | 0.573 (0.433) | 1.375** (0.477) | 0.426 (0.466) | 0.362* (0.178) |
| $\beta_0$ | -0.012* (0.005) | 0.001 (0.006) | -0.013* (0.006) | -0.011 (0.006) | -0.028***(0.002) |
| $\beta$ | 0.024* (0.009) | 0.015 (0.011) | 0.016 (0.012) | 0.020 (0.012) | 0.041***(0.005) |
| $R^2$ | 0.172 | 0.190 | 0.141 | 0.039 | 0.561 |
| Adj. $R^2$ | 0.153 | 0.171 | 0.120 | 0.016 | 0.550 |
| No. Observations | 132 | 132 | 132 | 132 | 132 |

Table 9: **OLS regression parameters for quote-level aggregation for Republicans shown in Figure 2 in the main text**. SEs of coefficients are in parantheses. ***$p < 0.001$, **$p < 0.01$ and *$p < 0.05$

|  | negative emotion | anger | anxiety | sadness | swear words |
| --- | --- | --- | --- | --- | --- |
| $\alpha_0$ | -1.417***(0.231) | -0.856** (0.256) | -1.543***(0.259) | -1.469***(0.255) | -1.345***(0.143) |
| $\alpha$ | 2.323***(0.372) | 1.990***(0.413) | 1.772***(0.418) | 1.437***(0.411) | 1.394***(0.231) |
| $\beta_0$ | -0.035***(0.005) | -0.022***(0.005) | -0.032***(0.006) | -0.034***(0.005) | -0.041***(0.003) |
| $\beta$ | 0.040***(0.010) | 0.026* (0.011) | 0.045***(0.011) | 0.018 (0.011) | 0.035***(0.006) |
| $R^2$ | 0.394 | 0.299 | 0.295 | 0.267 | 0.601 |
| Adj. $R^2$ | 0.380 | 0.282 | 0.278 | 0.249 | 0.592 |
| No. Observations | 132 | 132 | 132 | 132 | 132 |

Table 10: **OLS regression parameters for quote-level aggregation for Republicans, but excluding Donald Trump**. SEs of coefficients are in parantheses. ***$p < 0.001$, **$p < 0.01$ and *$p < 0.05$

|  | negative emotion | anger | anxiety | sadness | swear words |
| --- | --- | --- | --- | --- | --- |
| $\alpha_0$ | -1.456***(0.211) | -0.870***(0.237) | -1.548***(0.266) | -1.508***(0.244) | -1.369***(0.141) |
| $\alpha$ | 1.252***(0.340) | 1.156** (0.382) | 1.414** (0.429) | 0.979* (0.393) | 0.889***(0.227) |
| $\beta_0$ | -0.035***(0.005) | -0.022***(0.005) | -0.032***(0.006) | -0.034***(0.005) | -0.041***(0.003) |
| $\beta$ | 0.030***(0.009) | 0.019 (0.010) | 0.045***(0.011) | 0.015 (0.010) | 0.039***(0.006) |
| $R^2$ | 0.343 | 0.134 | 0.235 | 0.370 | 0.644 |
| Adj. $R^2$ | 0.328 | 0.113 | 0.217 | 0.355 | 0.636 |
| No. Observations | 132 | 132 | 132 | 132 | 132 |

Table 11: **OLS regression parameters for aggregation on the most prominent speaker quartile**. SEs of coefficients are in parantheses. ***$p < 0.001$, **$p < 0.01$ and *$p < 0.05$

|  | negative emotion | anger | anxiety | sadness | swear words |
| --- | --- | --- | --- | --- | --- |
| $\alpha_0$ | -2.661***(0.131) | -2.217***(0.121) | -1.673***(0.243) | -1.611***(0.244) | -1.288***(0.089) |
| $\alpha$ | 0.712***(0.210) | 0.601** (0.194) | 0.602 (0.391) | 0.323 (0.394) | 0.337* (0.143) |
| $\beta_0$ | -0.025***(0.003) | -0.008** (0.003) | -0.021***(0.005) | -0.018***(0.005) | -0.019***(0.002) |
| $\beta$ | 0.029***(0.005) | 0.011* (0.005) | 0.040***(0.010) | 0.031** (0.010) | 0.024***(0.004) |
| $R^2$ | 0.387 | 0.170 | 0.162 | 0.099 | 0.468 |
| Adj. $R^2$ | 0.373 | 0.150 | 0.143 | 0.078 | 0.456 |
| No. Observations | 132 | 132 | 132 | 132 | 132 |

Table 12: **OLS regression parameters for aggregation on the 2nd most prominent speaker quartile**. SEs of coefficients are in parantheses. ***$p < 0.001$, **$p < 0.01$ and *$p < 0.05$

|  | negative emotion | anger | anxiety | sadness | swear words |
| --- | --- | --- | --- | --- | --- |
| $\alpha_0$ | -4.272***(0.153) | -3.539***(0.141) | -2.721***(0.273) | -3.229***(0.398) | -1.534***(0.155) |
| $\alpha$ | 1.012***(0.247) | 0.599** (0.227) | 0.547 (0.440) | 0.417 (0.641) | 0.391 (0.250) |
| $\beta_0$ | -0.035***(0.003) | -0.013***(0.003) | -0.018** (0.006) | -0.030***(0.009) | -0.016***(0.003) |
| $\beta$ | 0.045***(0.006) | 0.018** (0.006) | 0.055***(0.011) | 0.062***(0.017) | 0.013 (0.006) |
| $R^2$ | 0.476 | 0.157 | 0.255 | 0.123 | 0.194 |
| Adj. $R^2$ | 0.464 | 0.137 | 0.238 | 0.102 | 0.175 |
| No. Observations | 132 | 132 | 132 | 132 | 132 |

Table 13: **OLS regression parameters for aggregation on the 3rd most prominent speaker quartile**. SEs of coefficients are in parantheses. ***$p < 0.001$, **$p < 0.01$ and *$p < 0.05$

|  | negative emotion | anger | anxiety | sadness | swear words |
|---|---|---|---|---|---|
| $\alpha_0$ | -4.703***(0.260) | -3.539***(0.225) | -2.937***(0.455) | -3.128***(0.705) | -1.397***(0.186) |
| $\alpha$ | 1.067* (0.419) | 0.331 (0.363) | 0.686 (0.733) | 0.734 (1.136) | 0.145 (0.300) |
| $\beta_0$ | -0.045***(0.006) | -0.013** (0.005) | -0.027** (0.010) | -0.043** (0.015) | -0.014***(0.004) |
| $\beta$ | 0.062***(0.011) | 0.026** (0.009) | 0.071***(0.019) | 0.057 (0.029) | 0.006 (0.008) |
| $R^2$ | 0.352 | 0.081 | 0.149 | 0.068 | 0.198 |
| Adj. $R^2$ | 0.337 | 0.060 | 0.129 | 0.046 | 0.179 |
| No. Observations | 132 | 132 | 132 | 132 | 132 |

Table 14: **OLS regression parameters for aggregation on the least prominent speaker quartile**. SEs of coefficients are in parantheses. ***$p < 0.001$, **$p < 0.01$ and *$p < 0.05$

|  | negative emotion | anger | anxiety | sadness | swear words |
|---|---|---|---|---|---|
| $\alpha_0$ | -3.425***(0.623) | -3.402***(0.600) | -2.429** (0.929) | -0.980 (1.738) | -1.042 (0.735) |
| $\alpha$ | -0.163 (1.005) | 0.493 (0.967) | -0.327 (1.497) | -1.859 (2.801) | 0.729 (1.185) |
| $\beta_0$ | -0.015 (0.013) | -0.017 (0.013) | -0.021 (0.020) | -0.004 (0.037) | -0.017 (0.016) |
| $\beta$ | 0.019 (0.026) | 0.024 (0.025) | 0.049 (0.039) | 0.044 (0.073) | 0.017 (0.031) |
| $R^2$ | 0.023 | 0.015 | 0.015 | 0.006 | 0.009 |
| Adj. $R^2$ | 0.000 | -0.008 | -0.008 | -0.017 | -0.014 |
| No. Observations | 132 | 132 | 132 | 132 | 132 |

Table 15: **OLS regression parameters for quote-level aggregation on speaker groups based on party affiliation, party's federal role, Congress membership, and gender**. SEs of coefficients are in parantheses. ***$p < 0.001$, **$p < 0.01$ and *$p < 0.05$

|  | negative emotion | anger | anxiety | sadness | swear words |
|---|---|---|---|---|---|
| $\alpha_0$ | -1.787***(0.183) | -1.520***(0.207) | -1.681***(0.203) | -1.198***(0.266) | -0.853***(0.100) |
| $\alpha$ | 1.293***(0.240) | 0.800** (0.271) | 1.960***(0.265) | 0.983** (0.347) | 0.522***(0.131) |
| $\beta_0$ | -0.013***(0.003) | 0.002 (0.004) | -0.018***(0.004) | -0.013** (0.005) | -0.033***(0.002) |
| $\beta$ | 0.014* (0.006) | 0.009 (0.007) | 0.006 (0.007) | -0.006 (0.009) | 0.043***(0.003) |
| $\gamma$: party affiliation | 0.479***(0.124) | 0.559***(0.140) | 0.964***(0.137) | 0.668***(0.179) | -0.193** (0.068) |
| $\eta$: gender | -0.215 (0.115) | -0.172 (0.130) | 0.480***(0.128) | 0.413 (0.167) | -0.828***(0.063) |
| $\zeta$: Congress membership | 3.412***(0.115) | 3.767***(0.130) | 1.734***(0.128) | 2.255***(0.167) | -0.150* (0.063) |
| $\delta$: party's federal role | -0.335** (0.124) | -0.185 (0.140) | -0.357** (0.137) | -0.976***(0.179) | -0.328***(0.068) |
| $R^2$ | 0.478 | 0.473 | 0.235 | 0.184 | 0.390 |
| Adj. $R^2$ | 0.475 | 0.470 | 0.230 | 0.178 | 0.386 |
| No. Observations | 1056 | 1056 | 1056 | 1056 | 1056 |

Table 16: **OLS regression parameters for speaker-level aggregation on speaker groups based on party affiliation, party's federal role, Congress membership, and gender**. SEs of coefficients are in parantheses. ***$p < 0.001$, **$p < 0.01$ and *$p < 0.05$

|  | negative emotion | anger | anxiety | sadness | swear words |
|---|---|---|---|---|---|
| $\alpha_0$ | -3.949***(0.176) | -3.361***(0.175) | -2.938***(0.301) | -2.619***(0.462) | -1.196***(0.185) |
| $\alpha$ | 0.734** (0.230) | 0.388 (0.229) | 1.058** (0.393) | 0.658 (0.603) | 0.223 (0.242) |
| $\beta_0$ | -0.023***(0.003) | -0.002 (0.003) | -0.024***(0.005) | -0.019* (0.008) | -0.021***(0.003) |
| $\beta$ | 0.030***(0.006) | 0.013* (0.006) | 0.038***(0.010) | 0.013 (0.016) | 0.026***(0.006) |
| $\gamma$: party affiliation | 0.904***(0.119) | 0.810***(0.118) | 0.983***(0.203) | 1.242***(0.312) | -0.002 (0.125) |
| $\eta$: gender | 0.528***(0.111) | 0.308** (0.110) | 1.070***(0.189) | 1.057***(0.290) | -0.761***(0.116) |
| $\zeta$: Congress membership | 3.101***(0.111) | 3.184***(0.110) | 1.742***(0.189) | 2.586***(0.290) | 0.264* (0.116) |
| $\delta$: party's federal role | -0.744***(0.119) | -0.544***(0.118) | -0.694***(0.203) | -1.219***(0.312) | -0.243 (0.125) |
| $R^2$ | 0.472 | 0.472 | 0.142 | 0.104 | 0.099 |
| Adj. $R^2$ | 0.469 | 0.469 | 0.137 | 0.098 | 0.093 |
| No. Observations | 1056 | 1056 | 1056 | 1056 | 1056 |

Table 17: **Most frequently quoted politicians and the Wikidata identifiers (QID)**.

|    | QID      | Name             | Number of quotes | Party      | Gender |
|----|----------|------------------|------------------|------------|--------|
| 1  | Q76      | Barack Obama     | 1499080          | Democrat   | M      |
| 2  | Q22686   | Donald Trump     | 762822           | Republican | M      |
| 3  | Q4496    | Mitt Romney      | 281774           | Republican | M      |
| 4  | Q6294    | Hillary Clinton  | 230306           | Democrat   | F      |
| 5  | Q207     | George W. Bush   | 200013           | Republican | M      |
| 6  | Q10390   | John McCain      | 160805           | Republican | M      |
| 7  | Q6279    | Joe Biden        | 127182           | Democrat   | M      |
| 8  | Q43144   | Sarah Palin      | 125980           | Republican | F      |
| 9  | Q170581  | Nancy Pelosi     | 101583           | Democrat   | F      |
| 10 | Q359442  | Bernie Sanders   | 96701            | Democrat   | M      |
| 11 | Q63879   | Chris Christie   | 96361            | Republican | M      |
| 12 | Q22316   | John Kerry       | 93036            | Democrat   | M      |
| 13 | Q203966  | Paul Ryan        | 92407            | Republican | M      |
| 14 | Q215057  | Rick Perry       | 89975            | Republican | M      |
| 15 | Q23505   | George H. W. Bush| 88859            | Republican | M      |
| 16 | Q11673   | Andrew Cuomo     | 88433            | Democrat   | M      |
| 17 | Q1124    | Bill Clinton     | 88173            | Democrat   | M      |
| 18 | Q182788  | Newt Gingrich    | 86670            | Republican | M      |
| 19 | Q355522  | Mitch McConnell  | 82061            | Republican | M      |
| 20 | Q11702   | John Boehner     | 80793            | Republican | M      |
| 21 | Q380900  | Chuck Schumer    | 75358            | Democrat   | M      |
| 22 | Q314459  | Harry Reid       | 73980            | Democrat   | M      |
| 23 | Q324546  | Marco Rubio      | 70349            | Republican | M      |
| 24 | Q434706  | Elizabeth Warren | 67699            | Democrat   | F      |
| 25 | Q22212   | Lindsey Graham   | 66895            | Republican | M      |
| 26 | Q439729  | Rick Scott       | 65523            | Republican | M      |
| 27 | Q13133   | Michelle Obama   | 65410            | Democrat   | F      |
| 28 | Q2036942 | Ted Cruz         | 64380            | Republican | M      |
| 29 | Q607     | Michael Bloomberg| 64366            | Democrat   | M      |
| 30 | Q553254  | Scott Walker     | 64326            | Republican | M      |



\end{document}

%% file: dlab_macros.tex
\usepackage[utf8]{inputenc}
\usepackage[T1]{fontenc}
\usepackage{hyphenat}
\usepackage{xspace}
\usepackage{amsmath}
\usepackage{amsfonts}
\usepackage[colorlinks=true,linkcolor=black,citecolor=black]{hyperref}
\usepackage{url}
\usepackage{booktabs}
\usepackage{multirow}
\usepackage{makecell}
\usepackage{caption}
\usepackage{minibox}
\usepackage{bbm}
\usepackage{graphicx}
\usepackage{balance}
\usepackage{mathtools}
\usepackage{color}
\usepackage{marvosym}
\usepackage{ifthen}
\usepackage{textcomp}
\usepackage{enumitem}
\usepackage{verbatim}
\usepackage{algorithm}
\usepackage{numprint}
\usepackage{balance}

\usepackage{amsthm}
\theoremstyle{plain}

\newcommand{\chatoDisplayMode}[1]{#1}

\definecolor{MyRed}{rgb}{0.6,0.0,0.0} 
\definecolor{MyBlack}{rgb}{0.1,0.1,0.1} 
\newcommand{\inred}[1]{{\color{MyRed}\sf\textbf{\textsc{#1}}}}
\newcommand{\frameit}[2]{
  \begin{center}
  {\color{MyRed}
  \framebox[.9\columnwidth][l]{
    \begin{minipage}{.85\columnwidth}
    \inred{#1}: {\sf\color{MyBlack}#2}
    \end{minipage}
  }\\
  }
  \end{center}
}

\newcommand{\note}[2][]{\chatoDisplayMode{\def\@tmpsig{#1}\frameit{{\Pointinghand} Note}{#2\ifx \@tmpsig \@empty \else \mbox{ --\em #1}\fi}}}
\newcommand{\todo}[2][]{\chatoDisplayMode{\def\@tmpsig{#1}\frameit{{\Writinghand} To-do}{#2\ifx \@tmpsig \@empty \else \mbox{ --\em #1}\fi}}}

\newcommand{\abbrevStyle}[1]{#1}
\newcommand{\ie}{\abbrevStyle{i.e.}\xspace}
\newcommand{\eg}{\abbrevStyle{e.g.}\xspace}
\newcommand{\cf}{\abbrevStyle{cf.}\xspace}
\newcommand{\etal}{\abbrevStyle{et al.}\xspace}
\newcommand{\vs}{\abbrevStyle{vs.}\xspace}

\newcommand{\Eqnref}[1]{Eq.~\ref{#1}}

\newcommand{\Tabref}[1]{Table~\ref{#1}}
\newcommand{\Figref}[1]{Fig.~\ref{#1}}

\newcommand{\xhdr}[1]{\vspace{1.7mm}\noindent{{\bf #1.}}}

\renewcommand{\textcite}[1]{\citeauthor{#1} \shortcite{#1}}

\newcommand{\hide}[1]{}

\makeatletter
\newcommand{\iffont}[2]{\ifthenelse{\equal{\f@family}{#1}}{#2}{}}
\makeatother

  \usepackage{mathptmx}

  \DeclareSymbolFont{greek}{OML}{cmm}{m}{n}
  \DeclareMathSymbol{\alpha}{\mathalpha}{greek}{"0B}
  \DeclareMathSymbol{\beta}{\mathalpha}{greek}{"0C}
  \DeclareMathSymbol{\gamma}{\mathalpha}{greek}{"0D}
  \DeclareMathSymbol{\delta}{\mathalpha}{greek}{"0E}
  \DeclareMathSymbol{\epsilon}{\mathalpha}{greek}{"0F}
  \DeclareMathSymbol{\zeta}{\mathalpha}{greek}{"10}
  \DeclareMathSymbol{\eta}{\mathalpha}{greek}{"11}
  \DeclareMathSymbol{\theta}{\mathalpha}{greek}{"12}
  \DeclareMathSymbol{\iota}{\mathalpha}{greek}{"13}
  \DeclareMathSymbol{\kappa}{\mathalpha}{greek}{"14}
  \DeclareMathSymbol{\lambda}{\mathalpha}{greek}{"15}
  \DeclareMathSymbol{\mu}{\mathalpha}{greek}{"16}
  \DeclareMathSymbol{\nu}{\mathalpha}{greek}{"17}
  \DeclareMathSymbol{\xi}{\mathalpha}{greek}{"18}
  \DeclareMathSymbol{\pi}{\mathalpha}{greek}{"19}
  \DeclareMathSymbol{\rho}{\mathalpha}{greek}{"1A}
  \DeclareMathSymbol{\sigma}{\mathalpha}{greek}{"1B}
  \DeclareMathSymbol{\tau}{\mathalpha}{greek}{"1C}
  \DeclareMathSymbol{\upsilon}{\mathalpha}{greek}{"1D}
  \DeclareMathSymbol{\phi}{\mathalpha}{greek}{"1E}
  \DeclareMathSymbol{\chi}{\mathalpha}{greek}{"1F}
  \DeclareMathSymbol{\psi}{\mathalpha}{greek}{"20}
  \DeclareMathSymbol{\omega}{\mathalpha}{greek}{"21}
  \DeclareMathSymbol{\varepsilon}{\mathalpha}{greek}{"22}
  \DeclareMathSymbol{\vartheta}{\mathalpha}{greek}{"23}
  \DeclareMathSymbol{\varpi}{\mathalpha}{greek}{"24}
  \DeclareMathSymbol{\varrho}{\mathalpha}{greek}{"25}
  \DeclareMathSymbol{\varsigma}{\mathalpha}{greek}{"26}
  \DeclareMathSymbol{\varphi}{\mathalpha}{greek}{"27}
  \DeclareSymbolFont{otone}{OT1}{cmr}{m}{n}
  \DeclareMathSymbol{\Gamma}{\mathalpha}{otone}{0}
  \DeclareMathSymbol{\Delta}{\mathalpha}{otone}{1}
  \DeclareMathSymbol{\Theta}{\mathalpha}{otone}{2}
  \DeclareMathSymbol{\Lambda}{\mathalpha}{otone}{3}
  \DeclareMathSymbol{\Xi}{\mathalpha}{otone}{4}
  \DeclareMathSymbol{\Pi}{\mathalpha}{otone}{5}
  \DeclareMathSymbol{\Sigma}{\mathalpha}{otone}{6}
  \DeclareMathSymbol{\Upsilon}{\mathalpha}{otone}{7}
  \DeclareMathSymbol{\Phi}{\mathalpha}{otone}{8}
  \DeclareMathSymbol{\Psi}{\mathalpha}{otone}{9}
  \DeclareMathSymbol{\Omega}{\mathalpha}{otone}{10}
  \DeclareSymbolFont{syms}{OML}{cmm}{m}{it}
  \DeclareMathSymbol{\partial}{\mathord}{syms}{"40}
  \DeclareMathAlphabet{\mathbold}{OML}{cmm}{b}{it}
  \DeclareSymbolFont{largesymbols}{OMX}{cmex}{m}{n}